\documentclass{article}

\PassOptionsToPackage{square, sort&compress, numbers}{natbib}

\usepackage[final]{neurips_2022}

\usepackage[utf8]{inputenc} 
\usepackage[T1]{fontenc}    
\usepackage{hyperref}       
\usepackage{url}            
\usepackage{booktabs}       
\usepackage{amsfonts}       
\usepackage{nicefrac}       
\usepackage{microtype}      
\usepackage{xcolor}         
\usepackage{floatrow}
\usepackage{blindtext}
\usepackage{mwe}
\usepackage{floatrow}
\usepackage{subfig}
\usepackage{wrapfig}
\usepackage{mathtools}
\usepackage{makecell}
\usepackage{hyperref}
\usepackage{siunitx}
\usepackage{bm}
\newcommand{\eq}{eq.~}
\newcommand{\eqs}{eqs.~}
\newcommand{\fig}{Fig.~}
\newcommand{\tab}{Tab.~}
\newcommand{\wrt}{w.\,r.\,t.~}
\newcommand{\cf}{cf.~}

\newcommand{\RN}[1]{\mathbb{R}^{#1}}
\newcommand{\lmax}{l_{\mathrm{max}}}
\newcommand{\lmin}{l_{\mathrm{min}}}
\newcommand{\sph}[2]{Y_{#1}^{#2}}
\newcommand{\newsph}[2]{\tilde{Y}_{#1}^{#2}}
\newcommand{\sphvec}[1]{\bm{Y}^{(#1)}}
\newcommand{\newsphvec}[1]{\bm{\tilde{Y}}^{(#1)}}
\newcommand{\br}{\bm{r}}
\newcommand{\nbr}{\hat{\bm{r}}} 

\newcommand{\mrot}{R_{\theta}}
\newcommand{\pce}{\mathcal{P}_{3D}}
\newcommand{\sphc}{\boldsymbol{\chi}}
\newcommand{\Sphc}{\mathcal{X}}
\newcommand{\neigh}[1]{\mathcal{N}(#1)}
\newcommand{\sneigh}[1]{\mathcal{N}_{\sphc}(#1)}

\newcommand{\an}{\si{\angstrom}}

\newcommand{\nlayer}{n_l}

\newcommand{\rcut}{r_{\mathrm{cut}}}
\newcommand{\frcut}{\phi_{\rcut}}
\newcommand{\fscut}{\phi_{\tilde{\chi}_{\rm cut}}}

\definecolor{oliver}{rgb}{0.1,0.6,0.1}

\makeatletter
\def\blfootnote{\xdef\@thefnmark{}\@footnotetext}
\makeatother

\title{So3krates: Equivariant attention for interactions on arbitrary length-scales in molecular systems}

\author{%
  J. Thorben Frank\textsuperscript{1,2}\thanks{thorbenjan.frank@googlemail.com}
  \And
  Oliver T. Unke\textsuperscript{1,2,3}
  \And
  Klaus-Robert Müller\textsuperscript{1,2,3,4,5}\thanks{klaus-robert.mueller@tu-berlin.de}\\\\
  \textsuperscript{1} Machine Learning Group, TU Berlin, 10587 Berlin, Germany\\
  \textsuperscript{2} BIFOLD, Berlin Institute for the Foundations of Learning and Data, Germany\\
  \textsuperscript{3} Google Research, Brain team, Berlin\\
  \textsuperscript{4} Department of Artificial Intelligence, Korea University, Seoul 136-713, Korea\\
  \textsuperscript{5} Max Planck Institut für Informatik, 66123 Saarbrücken, Germany
}

\begin{document}
\maketitle

\begin{abstract}
The application of machine learning methods in quantum chemistry has enabled the study of numerous chemical phenomena, which are computationally intractable with traditional \textit{ab-initio} methods. However, some quantum mechanical properties of molecules and materials depend on non-local electronic effects, which are often neglected due to the difficulty of modeling them efficiently. This work proposes a modified attention mechanism adapted to the underlying physics, which allows to recover the relevant non-local effects. Namely, we introduce \textit{spherical harmonic coordinates} (SPHCs) to reflect higher-order geometric information for each atom in a molecule, enabling a non-local formulation of attention in the SPHC space. Our proposed model \textsc{So3krates}\footnote{\url{https://github.com/thorben-frank/mlff}} -- a self-attention based message passing neural network -- uncouples geometric information from atomic features, making them independently amenable to attention mechanisms. Thereby we construct \textit{spherical filters}, which extend the concept of continuous filters in Euclidean space to SPHC space and serve as foundation for a \textit{spherical self-attention} mechanism. We show that in contrast to other published methods, \textsc{So3krates} is able to describe non-local quantum mechanical effects over arbitrary length scales. Further, we find evidence that the inclusion of higher-order geometric correlations increases data efficiency and improves generalization. \textsc{So3krates} matches or exceeds state-of-the-art performance on popular benchmarks, notably, requiring a significantly lower number of parameters (0.25--0.4x) while at the same time giving a substantial speedup (6--14x for training and 2--11x for inference) compared to other models.
\end{abstract}
\section{Introduction}
\begin{figure}
    \centering
    \includegraphics{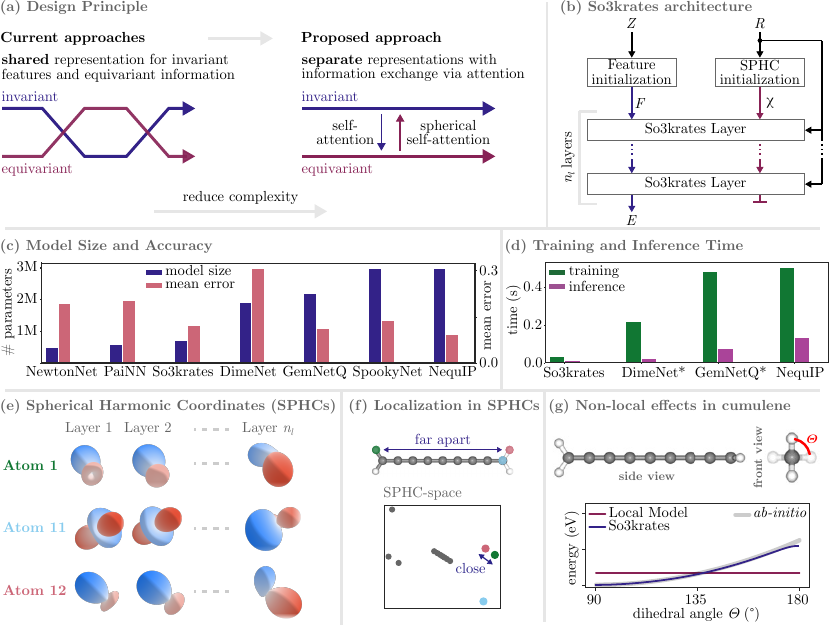}
    \caption{
    \textbf{(a)}~Design principle of our proposed message-passing scheme, where invariant features $\mathbf{f}$ and equivariant information - represented by \textit{spherical harmonic coordinates} (SPHCs) $\sphc$ - are separate and exchange information via (spherical) self-attention.
    \textbf{(b)}~Overview of the \textsc{So3krates} architecture.
    \textbf{(c)}~Comparison of prediction accuracy and model size of different architectures. \textbf{(d)}~Comparison of training and inference times of different architectures.
    \textbf{(e)}~Illustration of learned SPHCs at different layers of the \textsc{So3krates} architecture. \textbf{(f)}~Low-dimensional projection of atomic SPHCs, showing that atoms far apart in Euclidean space can be mapped close together in SPHC space. \textbf{(g)}~In contrast to local models, \textsc{So3krates} is able to learn the long-range correlations between hydrogen rotors in cumulene and reproduces the \textit{ab-initio} ground truth faithfully.
    }
    \label{fig:intro-figure}
\end{figure}

Atomistic simulations use long time-scale molecular dynamics (MD) trajectories to predict macroscopic properties that arise from interactions on the microscopic scale \cite{tuckerman2002ab,noe2020machine,keith2021combining}. Their predictive reliability is determined by the accuracy of the underlying \textit{force field} (FF), which needs to be queried at every time step. This quickly becomes a computational bottleneck if the forces are determined from first principles, which may be required for accurate results. To that end, machine learning FFs (MLFFs) offer a computationally more efficient, yet accurate empirical alternative to expensive \textit{ab-initio} methods \cite{behler2007generalized, bartok2010gaussian, behler2011atom, bartok2013representing, li2015molecular, chmiela2017machine, schutt2017quantum, gastegger2017machine, chmiela2018, schutt2018schnet, gilmer2017neural, smith2017ani, lubbers2018hierarchical, stohr2020accurate, faber2018alchemical, unke2019physnet, christensen2020fchl, unke2021machine, zhang2019embedded, kaser2020reactive, noe2020machine, von2020exploring}.

In recent years, \textit{Geometric Deep Learning} has become a popular design paradigm, which exploits relevant symmetry groups of the underlying learning problem by incorporating a \textit{geometric prior} \cite{chmiela2018,satorras2021n, bronstein2021geometric}. This effectively restricts the learnable functions of the model to a subspace with a meaningful inductive bias. Prominent examples for such models are e.g.\ convolutional neural networks (CNNs) \cite{lecun1995convolutional}, which are equivariant \wrt the group of translations, or graph neural networks (GNNs) \cite{kipf2016semi}, which are invariant \wrt node permutation.

For molecular property prediction, it has been shown that equivariance \wrt the 3D rotation group $\mathrm{SO}(3)$ greatly improves data efficiency and accuracy of the learned FFs \cite{klicpera2020directional,unke2021spookynet,schutt2021equivariant,batzner2021se}. To achieve equivariance, architectures either rely on feature expansions in terms of spherical harmonics (SH)~\cite{thomas2018tensor} or explicitly include (dihedral) angles \cite{klicpera2020directional, klicpera2021gemnet}. While the latter scales quadratically (cubically) in the number of neighboring atoms and has been shown to be geometrically incomplete~\cite{pozdnyakov2020incompleteness}, the calculation of spherical harmonics scales only linear in the number of neighboring atoms, which makes them a fast and accurate alternative \cite{batzner2021se, unke2021spookynet, fuchs2020se}.


\blfootnote{\textsuperscript{*}~Reference times were taken from \cite{klicpera2021gemnet}. As our own timings were measured on a different GPU, we decreased the reported times according to speedup-factors reported in \cite{AlishaAneja}. For full details, see appendix \ref{app:sec:runtime-analysis}.}

However, current higher-order geometric representations based on SHs usually result in expensive transformations, since an individual feature channel per SH degree (and order) is required. As a result, going to higher degrees is computationally expensive and comes at the price of increasing complexity, resulting in state-of-the-art (SOTA) models with millions of parameters \cite{batzner2021se, unke2021spookynet, klicpera2021gemnet}. However, in order to be applicable to large molecular structures, models are required to be both efficient and accurate on \emph{all} length scales. 

Non-local electronic effects have been outlined as one of the major challenges for a new generation of MLFFs \cite{unke2021machine}. They result in non-local, higher-order geometric relations between atoms. Most current architectures implicitly assume locality of interactions (expressed through a local neighborhood), which prohibits an efficient description of all relevant atomic interactions at larger scales. Simply increasing the cutoff radius used to determine local neighborhoods is not an adequate solution, since it only shifts the problem to larger length scales \cite{unke2021spookynet}.

In this work, we propose \textit{spherical harmonic coordinates} (SPHCs), which encode higher-order geometric information for each node in a molecular graph (Fig.~\ref{fig:intro-figure}e). This is in stark contrast to current approaches, which consider molecules as three-dimensional point clouds with learned features and fixed atomic coordinates: we propose to make the SPHCs themselves a {\em learned} quantity. Through localization in the space of SPHCs (Fig.~\ref{fig:intro-figure}f), models are able to efficiently describe electronic effects that are non-local in three-dimensional Euclidean space (Fig.~\ref{fig:intro-figure}g).

We then present \textsc{So3krates} (Fig.~\ref{fig:intro-figure}b), a self-attention based message-passing neural network (MPNN), which decouples atomic features and SPHCs and updates them individually (Fig.~\ref{fig:intro-figure}a). This resembles ideas from equivariant graph neural networks~\cite{satorras2021n}, but allows to go to arbitrarily high geometric orders. The separation of higher-order geometric and feature information allows to overcome the parametric and computational complexity usually encountered in models with higher-order geometric representations, since we only require a single feature channel (instead of one per SH degree and order). Thus, \textsc{So3krates} resembles some early architectures like \textsc{SchNet} \cite{schutt2017quantum} or \textsc{PhysNet} \cite{unke2019physnet} in parametric simplicity. We further show that \textsc{So3krates} outperforms the popular \textsc{sGDML} \cite{chmiela2019sgdml} kernel model by a large margin in the low-data regime, a domain which has so far been considered to be dominated by kernel machines \cite{unke2021machine}. Numerical evidence suggests that the data efficiency of \textsc{So3krates} is directly related to the maximal degree of geometric information encoded in the SPHCs. We then apply \textsc{So3krates} to the well-established MD17 benchmark and show that our model achieves SOTA results, despite is light-weight structure and having only 0.25--0.4x the number of parameters of competitive architectures (Fig.~\ref{fig:intro-figure}c), while achieving speedups of 6--14x and 2--11x for training and inference, respectively (Fig.~\ref{fig:intro-figure}d). 

Although we focus on quantum chemistry applications in this work, the developed methods are also applicable to other fields where long-ranged correlations in three-dimensional data are relevant. For example, models based on SPHCs may also be applicable to tasks like 3D shape classification or computer vision.

\section{Preliminaries and Related Work}

In the following, we review the most important concepts our method is based upon and relate it to prior work.

\paragraph{Message Passing Neural Networks}
MPNNs \cite{gilmer2017neural} carry over many of the benefits of convolutions to unstructured domains and have thus become one of the most promising approaches for the description of molecular properties. Their general working principle relies on the repeated iteration of message passing (MP) steps, which can be phrased as follows \cite{satorras2021n, gilmer2017neural}
\begin{align}
    \mathbf{m}_{ij} &= m(\mathbf{f}_i, \mathbf{f}_j, \br_{ij}) \label{eq:mp-message}\\
    \mathbf{m}_i &= \sum_{j \in \neigh{i}} \mathbf{m}_{ij} \\
    \mathbf{f}_i' &= u(\mathbf{f}_i, \mathbf{m}_i)\,.
\end{align}
Here, $\mathbf{m}_{ij}$ is the message between atoms $i$ and $j$ computed with the message function $m(\cdot)$, $\mathbf{m}_i$ is the aggregation of all messages in the neighborhood $\neigh{i}$ of atom~$i$, and $u(\cdot)$ is an update function returning updated features $\mathbf{f}_i'$ based on the current features $\mathbf{f}_i$ and message $\mathbf{m}_i$. The neighborhood $\neigh{i}$ consists of all atoms which lie within a given cutoff radius around the atomic position $\br_i$, which ensures linear scaling in the number of atoms~$n$. While earlier variants parametrized messages only in terms of inter-atomic distances \cite{schutt2018schnet, unke2019physnet}, more recent approaches also take higher-order geometric information into account \cite{anderson2019cormorant, klicpera2020directional,satorras2021n,schutt2021equivariant, batzner2021se,tholke2021equivariant}. 

\paragraph{Molecules as Point Clouds}
A molecule can be considered as a point cloud of $n$ atoms $\pce(\mathcal{R}, \mathcal{F})$, where $\mathcal{R}=(\br_1, ... \br_n)$ denotes the set of atomic positions $\br_i \in \RN{3}$ and $\mathcal{F}=(\mathbf{f}_1, \dots, \mathbf{f}_n)$ is the set of rotationally invariant atomic descriptors, or features, $\mathbf{f}_i \in \RN{F}$. We write the distance vector pointing from the position of atom~$i$ to the position of atom~$j$ as $\br_{ij} = \br_j - \br_i$, the distance as $r_{ij} = \lVert\br_{ij}\rVert_2$ and the normalized distance vector as $\hat{\br}_{ij} = \br_{ij} / r_{ij}$. Given the point cloud, a density  over Euclidean space assigning a vector value to each point $\br$ can be constructed as 
\begin{align}
    \boldsymbol{\rho}(\br) = \sum_{i=1}^n \delta(\lVert\br_i - \br\rVert_2) \cdot \mathbf{f}_i\,,
\end{align}
where $\delta$ is the Dirac delta function. It can be shown that applying a convolutional filter on $\boldsymbol{\rho}(\br)$ resembles the update steps used in MPNNs \cite{atzmon2018point}.

\paragraph{Equivariance} \label{sec:so3-equivariance}
Given a set of transformations that act on a vector space $\mathbb{A}$ as $S_g: \mathbb{A} \mapsto \mathbb{A}$ to which we associate an abstract group $G$, a function $f: \mathbb{A} \mapsto \mathbb{B}$ is said to be equivariant \wrt $G$ if
\begin{align}
    f(S_g x) = T_g f(x)\,, \label{eq:equivariance-definition}
\end{align}
where $T_g: \mathbb{B} \mapsto \mathbb{B}$ is an equivalent transformation on the output space \cite{satorras2021n}. Thus, in order to say that $f$ is equivariant, it must hold that under transformation of the input, the output transforms ``in the same way''. While equivariance has been a popular concept in signal processing for decades (\cf e.g.~\cite{cardoso1996equivariant} or wavelet neural networks~\cite{alexandridis2014wavelet}), recent years have seen efforts to design group equivariant NNs and kernel methods, since respecting relevant symmetries builds an important inductive bias \cite{cohen2016group, hinton2011transforming,chmiela2018}. Examples are CNNs \cite{lecun1995convolutional} which are equivariant \wrt translation, GNNs \cite{kipf2016semi, gilmer2017neural} which are invariant ($T_g = \mathbb{I}$) \wrt permutation, or architectures which are equivariant \wrt the $\mathrm{SO}(3)$ group \cite{thomas2018tensor,kohler2020equivariant,fuchs2020se,schutt2021equivariant,satorras2021n}. In this work, we consider the $\mathrm{SO}(3)$ group of rotations, such that $\mathbb{A}$ is the Euclidean space $\RN{3}$, where the corresponding group actions are given by rotation matrices $\mrot \in \RN{3 \times 3}$.

\paragraph{Spherical Harmonics} \label{sec:spherical-harmonics}
The spherical harmonics are special functions defined on the surface of the sphere $S^2 = \{\nbr \in \RN{3}: \lVert\nbr\rVert_2 = 1\}$ and form an orthonormal basis for the irreducible representations (irreps) of $\mathrm{SO}(3)$. In the context of \textit{tensor field networks} \cite{thomas2018tensor}, they have been introduced as elementary building blocks for $\mathrm{SO}(3)$-equivariant neural networks. The spherical harmonics are commonly denoted as $\sph{l}{m}(\nbr): S^2 \mapsto \RN{}$, where the \textit{degree} $l$ determines all possible values of the \textit{order} $m \in \{-l, \dots , +l\}$. They transform under rotation as
\begin{align}
    \sph{l}{m}(\mrot\,\nbr) = \sum_{m'} D_{mm'}^l(\mrot)\,\sph{l}{m'}(\nbr)\,,
    \label{eq:sph_equivariance}
\end{align}
where $D_{mm'}^l(\mrot)$ are the entries of the \textit{Wigner-D} matrix $\mathbf{D}^l(\mrot) \in \RN{(2l+1) \times (2l+1)}$ \cite{wigner1931gruppentheorie}.
Based on the spherical harmonics, we define a vector-valued function $\sphvec{l}: S^2 \mapsto \RN{2l+1}$ for each degree $l$, with entries $\sph{l}{m}$ for all valid orders $m$ of a given degree $l$. Since $\sphvec{l}(\mrot\,\nbr) = \mathbf{D}^l(\mrot)\sphvec{l}(\nbr)$ (\cf\eq\eqref{eq:sph_equivariance}), $\sphvec{l}$ is equivariant \wrt $\mathrm{SO}(3)$.

\paragraph{Tensor Product Contractions} \label{sec:tensor-contractions}
The irreps $\sphvec{l_1}$ and $\sphvec{l_2}$ can be coupled by computing their tensor product $\sphvec{l_1}\otimes\sphvec{l_2}$, which can equivalently be expressed as a direct sum \cite{thomas2018tensor,unke2021se}
\begin{align}
\sphvec{l_1}\otimes\sphvec{l_2} = \bigoplus_{l_3=\lvert l_1-l_2\rvert}^{l_1+l_2} \overbrace{\newsphvec{l_3}}^{\coloneqq\left(\sphvec{l_1}\otimes_{l_3}\sphvec{l_2}\right)}\,,
\label{eq:tensor-directsum}
\end{align} 
where the entry of order $m_3$ for the coupled irreps $\newsphvec{l_3}$ is given by
\begin{align}
    \newsph{m_3}{l_3} = \sum_{m_1=-l_1}^{l_1}\sum_{m_2 = -l_2}^{l_2} C_{m_1,m_2,m_3}^{l_1,l_2,l_3} \sph{m_1}{l_1} \sph{m_2}{l_2}\,, \label{eq:tensor-contraction}
\end{align}
and $C_{m_1,m_2,m_3}^{l_1,l_2,l_3}$ are the so-called \textit{Clebsch-Gordon coefficients}. In the following, we will denote the tensor product of degrees $l_1$ and $l_2$ followed by ``contraction'' to $l_3$ (meaning the irreps of degree $l_3$ in the direct sum representation of their tensor product) as $\left(\sphvec{l_1} \otimes_{l_3} \sphvec{l_2}\right)$, which is a mapping of the form $\RN{(2l_1+1) \times (2l_2+1)} \mapsto \RN{2l_3 +1}$, since $m_3 \in \{-l_3, \dots l_3\}$.

\section{Methods}

In the following, we describe the main methodological contributions of this work. We introduce the concept of an adapted point cloud $\pce(\mathcal{R}, \Sphc, \mathcal{F})$, which incorporates the set of spherical harmonics coordinates (SPHCs) $\Sphc=(\sphc_1, \dots, \sphc_n)$ (see below) in addition to features $\mathcal{F}$ and Euclidean coordinates $\mathcal{R}$. However, contrary to $\mathcal{R}$, SPHCs $\Sphc$ are refined during the message passing updates. Having SPHCs as part of the molecular point cloud extends the idea of current MPNNs, which learn message functions on $\mathcal{R}$, only. 
Instead, we learn a message function $m$ (\cf\eq\eqref{eq:mp-message}) on both, the (fixed) atomic coordinates $\mathcal{R}$ as well as on the SPHCs $\Sphc$. This adapted message-passing scheme allows to learn non-local geometric corrections. Based on these design principles, we propose the \textsc{So3krates} architecture. 

\paragraph{Initialization}
Feature vectors are initialized from the atomic numbers $z_i \in \mathbb{N}$ (denoting which chemical element an atom belongs to) by an embedding map 
\begin{align}
    \mathbf{f}_i = f_{\mathrm{emb}}(z_i), \label{eq:feature-initialization}
\end{align}
where $f_{\mathrm{emb}}: \mathbb{N} \mapsto \RN{F}$. We define SPHCs $\sphc$ as the concatenation of degrees $\mathcal{L} \coloneqq \{\lmin, \dots, \lmax\}$ 
\begin{align}
    \sphc = 
    [\underbrace{\sphc{^{(\lmin)}}}_{\in \RN{2\lmin+1}}, \dots, \underbrace{\sphc{^{(\lmax})}}_{\in \RN{2\lmax+1}}] \in \RN{(\lmax - \lmin + 1)^2},
\end{align}
such that their transformation under rotation can be expressed in terms of concatenated Wigner-D matrices (see appendix \ref{app:proof-of-equivariance}). The short-hand $\sphc^{(l)}\in\RN{2l+1}$ refers to the subset of SPHCs with degree~$l$. They are initialized as
\begin{align}
    \sphc_i^{(l)} =  \frac{1}{\mathcal{C}_i} \sum_{j \in \neigh{i}} \frcut(r_{ij}) \cdot \sphvec{l}(\nbr_{ij}), \label{eq:sphc-initialization}
\end{align}
where $\mathcal{C}_i = \sum_{j \in \neigh{i}} \frcut(r_{ij})$, $\frcut: \RN{} \mapsto \RN{}$ is the cosine cutoff function \cite{behler2011atom}, and the sum runs over the neighborhood $\mathcal{N}(i)$ of atom~$i$.

\paragraph{Message Passing Update} \label{sec:message-passing-update}
Two branches of attention-weighted MP steps are defined for the feature vectors $\mathbf{f}$ and SPHCs $\sphc$ (see \fig\ref{fig:intro-figure}a). After initialization (\eqs\eqref{eq:feature-initialization}~and~\eqref{eq:sphc-initialization}), the features are updated as
\begin{equation}
    \mathbf{f}_i' = \mathbf{f}_i + \sum_{j \in \neigh{i}} \frcut(r_{ij}) \cdot \alpha_{ij} \cdot \mathbf{f}_j\,, 
    \label{eq:feature-mp}
\end{equation}
where $\alpha_{ij} \in \RN{}$ are self-attention \cite{vaswani2017attention, velivckovic2017graph} coefficients (see below). In analogy to the feature vectors, it is possible to define an MP update for the SPHCs as
\begin{equation}
    \sphc_i^{\prime\,(l)} = \sphc_i^{(l)} + \sum_{j \in \neigh{i}} \frcut(r_{ij}) \cdot \alpha^{(l)}_{ij} \cdot  \sphvec{l}(\nbr_{ij})\,, \label{eq:sphc-mp}
\end{equation}
where individual attention coefficients  $\alpha_{ij}^{(l)} \in \RN{}$ for each degree of the SPHCs are computed using multi-head attention \cite{vaswani2017attention}. 
However, with this definition, both MP updates are limited to local neighborhoods $\mathcal{N}(i)$. To be able to model non-local effects, we introduce the SPHC distance matrix $\mathbf{X} \in \RN{n\times n}$ with entries $\chi_{ij} = \lVert\sphc_i - \sphc_j\rVert_2$, i.e.\ distances between two atoms~$i$~and~$j$ in SPHC space for all possible pair-wise combinations of $n$ atoms. To have uniform scales, we further apply the $\mathrm{softmax}$ along each row of $\mathbf{X}$ to generate a rescaled matrix $\tilde{\mathbf{X}} = \mathrm{softmax}(\mathbf{X})$ with entries $\tilde{\chi}_{ij}$. A polynomial cutoff function $\fscut$~\cite{klicpera2020directional} is then applied to $\tilde{\mathbf{X}}$ 
to define spherical neighborhoods $\sneigh{i}$ (see \ref{app:spherical-neighborhood}), which may include atoms that are far away in Euclidean space (see Fig.~\ref{fig:intro-figure}f). 
The spherical cutoff distance is chosen as $\tilde{\chi}_{\rm cut} = 1/n$ to ensure that  spherical neighborhoods remain small, even when going to larger molecules. We then incorporate non-local geometric corrections into the MP update of the SPHCs as
\begin{align}
    {\sphc}_i^{\prime\,(l)} = \sphc_i^{(l)} + \underbrace{\sum_{j \in \neigh{i}} \frcut(r_{ij}) \cdot \alpha_{ij}^{(l)} \cdot \sphvec{l}(\nbr_{ij})}_{\text{local in $\RN{3}$}} + \underbrace{\sum_{j \in \sneigh{i}} \fscut(\tilde{\chi}_{ij}) \cdot \alpha_{ij}^{(l)} \cdot \sphvec{l}(\nbr_{ij})}_{\text{local in ${\sphc}$, but non-local in $\RN{3}$}}\,.
    \label{eq:sphc-non-local-mp}
\end{align}
We will show in the first part of the experiments, how geometric corrections from SPHC space allow for modelling non-local quantum effects, inaccessible to current architectures. In the second part, we use a \textsc{So3krates} model without geometric corrections, which makes it a traditional MPNN in the sense of only localizing in $\RN{3}$. We find this architecture to be highly parameter, data and time efficient while capable of reaching SOTA results.


\paragraph{Spherical Filter and Self-Attention} \label{sec:self-attention}
The self-attention coefficients in \eqs\eqref{eq:feature-mp}--\eqref{eq:sphc-non-local-mp} are calculated as
\begin{align}
    \alpha_{ij} = \mathbf{f}_i^T (\mathbf{w}_{ij} \odot \mathbf{f}_j)/{\sqrt{F}}\,, 
    \label{eq:self-attention}
\end{align}
where $\mathbf{w}_{ij} \in \RN{F}$ is the output of a filter generating function and `$\odot$' denotes the element-wise product. The filter maps the Euclidean distance $r_{ij}$ and per-degree SPHC distances $\chi^{(l)}_{ij} = \lVert\sphc^{(l)}_{j} - \sphc^{(l)}_{i}\rVert_2$ between the current SPHCs of atoms~$i$~and~$j$ into the feature space $\RN{F}$ (as a short-hand, we write the vector containing all per-degree SPHC distances as $[\chi^{(l)}_{ij}]_{l\in\mathcal{L}}$). 
It is built as the linear combination of two filter-generating functions 
\begin{align}
    \mathbf{w}_{ij} = \underbrace{\phi_r(r_{ij})}_{\text{radial filter}} + \underbrace{\phi_s\left([\chi^{(l)}_{ij}]_{l\in\mathcal{L}}\right)}_{\text{spherical filter}}, \label{eq:filter}
\end{align}
which separately act on the Euclidean and SPHC distances. We call $\phi_r: \RN{} \mapsto \RN{F}$ the \textit{radial filter} function and $\phi_s: \RN{\lvert\mathcal{L}\rvert} \mapsto \RN{F}$ the \textit{spherical filter} function (an ablation study for $\phi_s$ can be found in appendix \ref{app:sec:ablation-study-spherical-filter}). Since per-atom features $\mathbf{f}_i$, interatomic distances $r_{ij}$, and per-degree distances $\chi^{(l)}_{ij}$ are invariant under rotations (proof in appendix \ref{app:proof-of-equivariance}), so are the self-attention coefficients $\alpha_{ij}$. 

While we choose to pass the per-degree norms directly into the filter generating function $\phi_s$, future work might explore the possibilities of alternative metrics (instead of the L2 norm) or an expansion in terms of basis functions as it is common practice for inter-atomic distances (see appendix \ref{app:network} \eq\eqref{eq:basis-fn-expansion}).

\paragraph{Atomwise Interaction} \label{sec:interaction}
After each MP update, features and SPHCs are coupled with each other according to
\begin{align}
    \mathbf{f}_i' &= \mathbf{f}_i + \phi_1\left(\mathbf{f}_i, [\chi_i^{(l)}]_{l\in\mathcal{L}}, [\tilde{\chi}_i^{(l)}]_{l\in\mathcal{L}}\right)\,, \label{eq:feature-interaction}\\
    \sphc_i^{\prime\,(l)} &= \sphc_i^{(l)} + \phi_2^{(l)}\left(\mathbf{f}_i, [\chi_i^{(l)}]_{l\in\mathcal{L}}, [\tilde{\chi}_i^{(l)}]_{l\in\mathcal{L}}\right) \sphc_i^{(l)}  + \phi_3^{(l)}\left([\tilde{\chi}_i^{(l)}]_{l\in\mathcal{L}}\right)\tilde{\sphc}_i^{(l)}\,,
    \label{eq:sphc-interaction}
\end{align}
where $\phi_1: \RN{F + 2\lvert\mathcal{L}\rvert} \mapsto \RN{F}$, $\phi^{(l)}_{2}: \RN{F + 2\lvert\mathcal{L}\rvert} \mapsto \RN{}$, and $\phi^{(l)}_{3}: \RN{\lvert\mathcal{L}\rvert} \mapsto \RN{}$. In the inputs to $\phi_{1,2,3}$, degree-wise scalars $\chi^{(l)} = \lVert\sphc^{(l)}\rVert_2$ are used to preserve equivariance. The coupling step additionally includes cross-degree coupled SPHCs $\tilde{\sphc}_i^{(l)}$ for each degree $l$. Following \cite{unke2021se} they are constructed as
\begin{align}
    \tilde{\sphc}_i^{(l)} = \sum_{l_1=\lmin}^{\lmax} \sum_{l_2=l_1 + 1}^{\lmax} k_{l_1, l_2, l} \,  \left(\sphc_i^{(l_1)} \otimes_l \sphc_i^{(l_2)}\right),
\end{align}
where $k_{l_1, l_2, l} \in \RN{}$ are learnable coefficients for all valid combinations of $l_1, l_2$ given $l$ and the term in brackets is the contraction of degrees $l_1$ and $l_2$ into degree $l$ (\eq\eqref{eq:tensor-contraction}).

\paragraph{\textsc{So3krates} architecture}
Using the design paradigm above, we build the transformer network \textsc{So3krates}, which consists of a self-attention block on $\mathcal{F}$ and $\Sphc$ (\eqs~\eqref{eq:feature-mp}~and~\eqref{eq:sphc-mp}), respectively, as well as an interaction block (\eqs~\eqref{eq:feature-interaction}~and~\eqref{eq:sphc-interaction}) per layer. After initialization of the features and the SPHCs according to \eqs\eqref{eq:feature-initialization}~and~\eqref{eq:sphc-initialization}, they are updated iteratively by passing through $\nlayer$ layers. Atomic energy contributions $E_i \in \RN{}$ are predicted from the features of the final layer using a two-layered output block. The individual contributions are summed to the total energy prediction $E = \sum_{i}^n E_i$. See Fig.~\ref{fig:intro-figure}b for an overview. More details on the implementation, training details and network hyperparameters are given in appendix \ref{app:network} and \ref{app:sec:training-details}.
\section{Experiments}
In the first subsection, we show how non-local quantum effects can be incorporated by using non-local corrections from the space of SPHCs. In the second part of the experiments, we remove the non-local part which yields a traditional, $\RN{3}$-local MPNN which reaches SOTA results on established benchmarks while requiring much less computational time and parameters than competitive models. A scaling analysis as well as an accuracy comparison for both model variants can be found in appendix \ref{app:sec:scaling} and \ref{app:sec:localvs-non-local-acc}.



\begin{figure}
    \centering
    \includegraphics{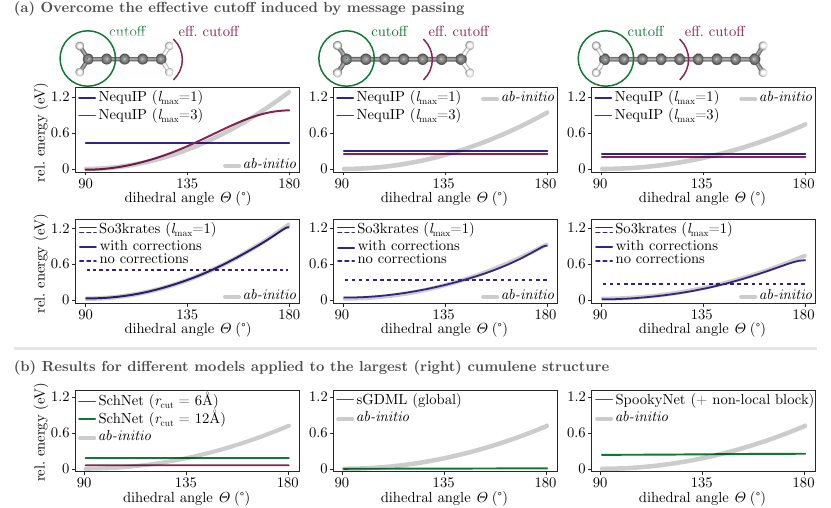}
    \caption{\textbf{(a)}~Energy predictions for different cumulene structures with \textsc{NequIP} ($\lmax =1$ and $\lmax = 3$) and \textsc{So3krates} ($\lmax =1$) models with and without geometric corrections (\eqs\eqref{eq:sphc-mp}~and~\eqref{eq:sphc-non-local-mp}). Cutoff radius and number of layers are kept constant at $\rcut = 2.5\,\an$ and $\nlayer = 4$, respectively. \textbf{(b)}~We additionally test other SOTA models on the largest cumulene structure (9 carbons). We find, that even for a very large cutoff, \textsc{SchNet} fails to describe the dihedral angle profile. The same holds true for an inherently global \textsc{sGDML} model and even for a \textsc{SpookyNet} model that explicitly includes non-local corrections.
    }
    \label{fig:non-local-experiment}
\end{figure}

\paragraph{Non-Local Geometric Interactions} \label{sec:non-local-geometric-interactions}
For efficiency reasons, MPNNs only consider interactions between atoms in local neighborhoods, i.e.\ within a cutoff radius $\rcut$. Thus, information can only be propagated over a distance of $\rcut$ within a single MP step. Although multiple MP updates increase the effective cutoff distance, because information can ``hop'' between different neighborhoods as long as they share at least one atom,
each MP step is accompanied by an undesirable loss of information, which limits the accuracy that can be obtained. Consequently, MPNNs are unable to describe non-local effects on length-scales that exceed the effective cutoff distance. To illustrate this problem, we consider the challenging open task \cite{unke2021machine} of learning the potential energy of cumulene molecules with different sizes (see \fig\ref{fig:non-local-experiment}a). Here, the relative orientation of the hydrogen rotors at the far ends of the molecule strongly influences its energy due to non-local electronic effects \cite{unke2021machine}. In order to be able to successfully learn the energy profile with a local model, the effective cutoff has to be large enough to allow information to propagate from one hydrogen rotor to the other.

As a representative example for MPNNs, we consider the recently proposed \textsc{NequIP} model~\cite{batzner2021se}, which achieves SOTA performance on several benchmarks. We find that even when the effective cutoff radius is large enough in principle, an MPNN with $\nlayer = 4$, $\rcut = 2.5\,\an$, and $\lmax \leq 1$ fails to learn the correct energy profile. This is due to the fact that the relevant geometric information ``cancels out'' (similar to addition of vectors oriented in opposite directions) within each neighborhood, underlining the limited expressiveness of mean-field interactions in MPNNs. Only by including higher-order geometric correlations, e.g.\ going to $\lmax = 3$, the correct energy profile can be recovered (at the cost of computational efficiency). When going to even larger cumulene structures, however, the effective cutoff becomes too small and it is necessary to increase the number of MP layers to solve the task (again, at the cost of lower computational efficiency), which is illustrated in appendix \ref{app:sec:additional-experiments-non-local-effects}. Neither increasing the maximum degree of interactions $\lmax$, nor the number of layers $\nlayer$, is a satisfactory workaround: Instead of offering a general solution to describe non-local interactions, both options decrease computational efficiency, while only shifting the problem to larger length-scales or higher-order geometric correlations.

We further apply three additional models to the cumulene structure with nine carbon atoms. To that end, we use an invariant \textsc{SchNet} model with varying cutoff distances (6\,$\si{\angstrom}$ and 12\,$\si{\angstrom}$), an inherently global but invariant \textsc{sGDML} model and the \textsc{SpookyNet} architecture which explicitly includes global effects using a non-local block. We find that none of the three is capable of describing the rotor energy profile of cumulene.

\begin{figure}
    \centering
    \includegraphics{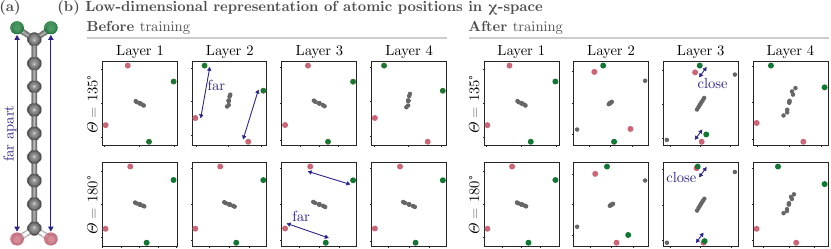}
    \caption{\textbf{(a)}~Cumulene structure in Euclidean space. \textbf{(b)}~Low-dimensional projections of the SPHCs ($\lmax = 1$) in \textsc{So3krates} before and after training. Carbon atoms are grey and hydrogen atoms on different ends of the molecule are green and red, respectively. During training, the model maps hydrogen atoms at opposite ends close together in SPHC space (which is not the case at initialization).}
    \label{fig:sphc-embedding}
\end{figure}

In contrast, our proposed \textsc{So3krates} architecture is able to reproduce the energy profile for cumulene molecules of all sizes independent of the effective cutoff radius. Crucially, even with $\lmax=1$, the predicted energy matches the \textit{ab-initio} reference faithfully. We find that geometric corrections in the MP update of the SPHCs (\cf\eq\eqref{eq:sphc-non-local-mp}) are responsible for the increased capability of describing higher-order geometric correlations, as a \textsc{So3krates} model with a naive MP update (\cf\eq\eqref{eq:sphc-mp}) fails  to solve this task with $\lmax=1$ (see \fig\ref{fig:non-local-experiment}a). We further confirm that the model picks up on the physically relevant interaction between the hydrogen rotors by analysing the attention values after training (see \fig\ref{fig:attention-analysis} appendix \ref{app:analysis-attention}). To illustrate how
\textsc{So3krates} is able to describe non-local effects, we show a low-dimensional projection of the atomic SPHCs before and after training for the largest of the cumulene molecules (\fig\ref{fig:sphc-embedding}). After training, the SPHCs for hydrogen atoms at opposite ends of the molecule are embedded close together in SPHC space, allowing \textsc{So3krates} to efficiently model the non-local geometric dependence between the hydrogen rotors.

Generalization to structures, larger than those in the training data are usually associated with the re-usability of the learned, local representations. For that reason, it is unclear if this property still holds when non-local corrections are used. As we show in appendix \ref{sec:transfer-leraning} a \textsc{So3krates} model with non-local corrections still generalizes well to completely unknown and larger structures.

\paragraph{Benchmarks, Data Efficiency and Generalization} \label{sec:benchmarks-data-efficiency-generalization}
As pointed out in \cite{schutt2021equivariant} and \cite{batzner2021se}, equivariant features not only increase performance, but also improve data efficiency. The latter is particularly important, as \textit{ab-initio} methods for reference data generation can become exceedingly expensive when high accuracy is required. Here, we use a subset of the recently introduced QM7-X data set \cite{hoja2021qm7}, which we call QM7-X250. It contains 250 different molecular structures, each with 80 data points for training, 10 data points for validation and 11-3748 data points for testing (for details, see appendix \ref{app:qm7-x250}). The small number of training/validation samples per molecule makes it particularly suited for evaluating model behavior in the low data regime. In the following, we train (1) one model per structure in QM7-X250 and (2) one model for all structures in QM7-X250 ($250 \times 80 = 20$k training points), which we refer to as \textit{individual} and \textit{joint} models, respectively.
\begin{figure}
    \centering
    \includegraphics{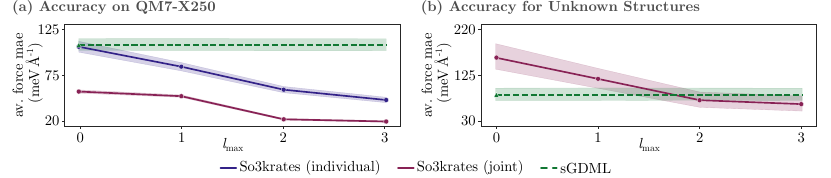}
    \caption{\textbf{(a)}~Average force MAE as a function of $\lmax$ on 250 structures from QM7-X~\cite{hoja2021qm7}. The blue solid line corresponds to \textsc{So3krates} models trained on each of the 250 structures (\`a 80 training points), individually. The jointly trained \textsc{So3krates} model (red solid line) is a single model trained on all $250 \times 80$ training points. \textbf{(b)}~Accuracy of force predictions for unknown structures as function of $\lmax$. Generalization is only investigated for the jointly trained model. Due to the way the molecular descriptor is designed, \textsc{sGDML}~\cite{chmiela2019sgdml} models can only be trained on individual structures.}
    \label{fig:qm7x}
\end{figure}

We start by investigating the performance as a function of the maximal degree $\lmax$ and find that the error strongly decreases with higher $\lmax$ (\fig\ref{fig:qm7x}a). As kernel methods are known to perform well in the low-data regime~\cite{unke2021machine}, we compare our results to \textsc{sGDML} \cite{chmiela2019sgdml} kernel models, which only use distances as a molecular descriptor (corresponding to $\lmax = 0$). For $\lmax=0$, we find \textsc{sGDML} gives competitive results, whereas for $\lmax=1$, \textsc{So3krates} starts to outperform \textsc{sGDML}. As soon as $\lmax\geq2$, however, the prediction accuracy of \textsc{So3krates} is greater than that of \textsc{sGDML} by a large margin. Thus, increasing the order of geometric information in the SPHCs leads to strong improvements in the low-data regime. For jointly trained models, we find that \textsc{So3krates} outperforms \textsc{sGDML} even for $\lmax = 0$, with continuous improvement for increasing $\lmax$. In appendix \ref{app:sec:qm7x-250-experiments} we report energy \textit{and} force errors across degrees and further experimental details. 
        
The generalization capability of \textsc{So3krates} is tested, by applying a jointly trained model to 25 completely unknown molecules from the QM7-X data set (see, Fig.~\ref{fig:qm7x}b, details in appendix \ref{app:sec:qm7x-250-experiments}). Again, we find that force MAEs decrease with increasing $\lmax$. For reference, we compare \textsc{So3krates} to individually trained \textsc{sGDML} models and find that \textsc{So3krates} performs on par, or even slightly better, for $\lmax \geq 2$. Going beyond $\lmax=2$ is found to only marginally improve generalization. In addition, we report results for a model trained on the full QM7-X data set in appendix \ref{app:sec:qm7x}, following \cite{unke2021spookynet}.

For completeness, we also apply \textsc{So3krates} to the popular MD17 benchmark (see Table~\ref{tab:MD17}). 
We find, that \textsc{So3krates} outperforms networks that have the same parameter complexity by a large margin (\textsc{PaiNN} and \textsc{NewtonNet}). Notably, it requires significantly less parameters than other SH based architectures (\textsc{NequIP} and \textsc{SpookyNet}), while performing only slightly worse or even on par with them. Furthermore, \textsc{So3krates} outperforms \textsc{DimeNet}, its closest competitor in timing (\cf\fig\ref{fig:intro-figure}.d), consistently by a large margin. Compared to current SH based approaches, \textsc{GemNetQ} needs less parameters (still $\sim 2.5$x more than \textsc{So3krates}) to achieve competitive results. However, it requires the explicit calculation of dihedral angles which scales cubically in the number of neighboring atoms. Due to its linear scaling (see \ref{app:sec:scaling}) and lightweight structure, \textsc{So3krates} can significantly reduce the time for training and inference (see \fig\ref{fig:intro-figure}d and \ref{app:sec:runtime-analysis}).
\begin{table*}[]
    \centering
    \resizebox{\linewidth}{!}{
    \begin{tabular}{c*{7}{c}|c}
    \toprule
         & & \thead{\textsc{NequIP}~\cite{batzner2021se} \\  3M | $\mathcal{O}(m)$} &  \thead{\textsc{SpookyNet}~\cite{unke2021spookynet} \\ 3M | $\mathcal{O}(m)$} &
         \thead{\textsc{GemNetQ}~\cite{klicpera2021gemnet} \\  2.2M | $\mathcal{O}(m^3)$}& \thead{\textsc{DimeNet}~\cite{klicpera2020directional} \\ 1.9M | $\mathcal{O}(m^2)$} & \thead{\textsc{PaiNN}~\cite{schutt2021equivariant} \\ 600k | $\mathcal{O}(m)$} &
         \thead{\textsc{NewtonNet}~\cite{haghighatlari2021newtonnet} \\ 500k | $\mathcal{O}(m)$} & \thead{\textsc{So3krates} \\ 700k | $\mathcal{O}(m)$} \\\midrule
         
         Aspirin & \thead{\textit{energy}\\\textit{forces}} & \thead{0.13\\0.19}  &\thead{0.151\\0.258}& \thead{--\\0.217} & \thead{0.204\\ 0.499} & \thead{0.159\\0.371} & \thead{0.168\\0.348} & \thead{0.139\\0.236}\\\midrule
         
         Ethanol & \thead{\textit{energy}\\\textit{forces}} & \thead{0.05\\0.09} & \thead{0.052\\0.094} & \thead{--\\0.088} & \thead{0.064\\0.230} & \thead{0.063\\0.230} & \thead{0.061\\0.211} &  \thead{0.052\\0.096}\\\midrule
         
         Malondialdehyde & \thead{\textit{energy}\\\textit{forces}} & \thead{0.08\\0.13} & \thead{0.079\\0.167} & \thead{--\\0.159} & \thead{0.104\\0.383} & \thead{0.091\\0.319} & \thead{0.096\\0.323} & \thead{0.077\\0.147}\\\midrule
         
         Naphthalene & \thead{\textit{energy}\\\textit{forces}} & \thead{0.11\\0.04} & \thead{0.116\\0.089} & \thead{--\\0.051} &  \thead{0.122\\0.215} & \thead{0.117\\0.083} & \thead{0.118\\0.084} &  \thead{0.115\\0.074}\\\midrule
         
         Salicyclic Acid & \thead{\textit{energy}\\\textit{forces}} & \thead{0.11\\0.09} & \thead{0.114\\0.180} & \thead{--\\0.124} & \thead{0.134\\0.374} & \thead{0.114\\0.209} & \thead{0.115\\0.197} & \thead{0.106\\0.145}\\\midrule
         
         Toluene & \thead{\textit{energy}\\\textit{forces}} & \thead{0.09\\0.05} & \thead{0.094\\0.087} & \thead{--\\0.060} & \thead{0.102\\0.216} & \thead{0.097\\0.102} & \thead{0.094\\0.088} &   \thead{0.095\\0.073} \\\midrule
         
         Uracil & \thead{\textit{energy}\\\textit{forces}} & \thead{0.10\\0.08} & \thead{0.105\\0.119} & \thead{--\\0.104} & \thead{0.115\\0.301} & \thead{0.104\\0.140} & \thead{0.107\\0.149} & \thead{0.103\\0.111}
         \\\bottomrule
    \end{tabular}
    } 
        \caption{MAE for energy (in kcal\,mol$^{-1}$) and forces (in kcal\,mol$^{-1}\,\si{\angstrom}^{-1}$) of current state-of-the-art machine learning models on the MD17 benchmark for 1k training points. For each model, the number of paramaeters, as well as the scaling in the number of neighbors $m$ is shown.
        }
    \label{tab:MD17}
\end{table*}

\section{Discussion and Conclusion}
Due to the locality assumption used in most MPNNs, they are unable to model non-local electronic effects, which result in global geometric dependencies between different parts of a molecule. The length-scales of such interactions often greatly exceed the cutoff radius used in the MP step, and even though stacking multiple MP layers increases the effective cutoff, ultimately, MPNNs are not capable of efficiently modeling geometric dependencies on arbitrary length scales.

In this work, we contribute conceptually by proposing an efficient and scalable solution to this problem. 
We suggest a set of refinable, equivariant coordinates for point clouds in Euclidean space, called spherical harmonic coordinates (SPHCs). Non-local geometric effects can then be efficiently modeled by including geometric corrections, which are localized in the space of SPHCs, but non-local in Euclidean space. Further, we show that introducing spherical filter functions acting on the SPHCs increases geometric resolution and predictive accuracy. 

We then propose the \textsc{So3krates} architecture, a self-attention based MPNN, which decouples  atomic features from higher-order geometric information. This allows to drastically decrease the parametric complexity while still achieving SOTA prediction accuracy. We show evidence that increasing the geometric order of SPHCs greatly improves model performance in the low-data regime, as well as generalization to unknown molecules. 

A limitation of the current implementation of \textsc{So3krates} is that spherical neighborhoods $\mathcal{N}_{\sphc}$ in \eq\eqref{eq:sphc-non-local-mp} are computed from all pairwise distances in SPHC space. An alternative implementation could use a space partitioning scheme to find neighborhoods more efficiently. In a broader context, our work falls into the category of approaches that can help to reduce the vast computational complexity of molecular and material simulations. This can accelerate novel drug and material designs, which holds the promise of tackling societal challenges, such as climate change and sustainable energy supply~\cite{jain2013commentary}. Of course, our method could also be used for nefarious applications, e.g.\ design of
chemical warfare, but this is true for all quantum chemistry methods.

Future research will focus on applications of \textsc{So3krates} to materials and bio-molecules, which are typical examples of chemical systems where the accurate description of non-local effects is necessary to produce novel insights.
Efficient treatment of non-local effects in point cloud data goes beyond the domain of quantum chemistry. One way of representing non-local dependencies are non-local neural networks~\cite{wang2018non}. In comparison to the presented approach they compute a relation in feature rather than in Euclidean space, making it incapable of capturing direct geometric relations in Euclidean space. However, this might be necessary if the relative orientation of objects far apart from each other plays a role for identifying different objects.

\section{Acknowledgements}
All authors acknowledge  support by the Federal Ministry of Education and Research (BMBF) for BIFOLD (01IS18037A). KRM was partly supported by the Institute of Information \& Communications Technology Planning \& Evaluation (IITP) grants funded by the Korea government(MSIT) (No. 2019-0-00079, Artificial Intelligence Graduate School Program, Korea University and No. 2022-0-00984, Development of Artificial Intelligence Technology for Personalized Plug-and-Play Explanation and Verification of Explanation), and was partly supported by the German Ministry for Education and Research (BMBF) under Grants 01IS14013A-E, AIMM, 01GQ1115, 01GQ0850, 01IS18025A and 01IS18037A; the German Research Foundation (DFG).
We thank Stefan Chmiela, Mihail Bogojeski and Nicklas Schmitz for helpful discussions and feedback on the manuscript.

%
\bibliographystyle{unsrtnat}
\bibliography{Bibliography}
\newpage
\begin{enumerate}

\item For all authors...
\begin{enumerate}
  \item Do the main claims made in the abstract and introduction accurately reflect the paper's contributions and scope?
    \answerYes{}
  \item Did you describe the limitations of your work?
    \answerYes{}
  \item Did you discuss any potential negative societal impacts of your work?
    \answerYes{}
  \item Have you read the ethics review guidelines and ensured that your paper conforms to them?
    \answerYes{}
\end{enumerate}

\item If you are including theoretical results...
\begin{enumerate}
  \item Did you state the full set of assumptions of all theoretical results?
    \answerNA{We did not include theoretical results.}
        \item Did you include complete proofs of all theoretical results?
    \answerNA{Even though we proof equivariance, this is not a theoretical result of the presented work, as equivariance \wrt $\mathrm{SO(3)}$ is well understood.}
\end{enumerate}

\item If you ran experiments...
\begin{enumerate}
  \item Did you include the code, data, and instructions needed to reproduce the main experimental results (either in the supplemental material or as a URL)?
    \answerYes{}
  \item Did you specify all the training details (e.g., data splits, hyperparameters, how they were chosen)?
    \answerYes{}
        \item Did you report error bars (e.g., with respect to the random seed after running experiments multiple times)?
    \answerYes{See \fig\ref{fig:qm7x} or \fig\ref{fig:sph_filter_ablation}.}
        \item Did you include the total amount of compute and the type of resources used (e.g., type of GPUs, internal cluster, or cloud provider)?
    \answerYes{}
\end{enumerate}

\item If you are using existing assets (e.g., code, data, models) or curating/releasing new assets...
\begin{enumerate}
  \item If your work uses existing assets, did you cite the creators?
    \answerYes{}
  \item Did you mention the license of the assets?
    \answerYes{}
  \item Did you include any new assets either in the supplemental material or as a URL?
    \answerYes{}
  \item Did you discuss whether and how consent was obtained from people whose data you're using/curating?
    \answerNA{No such data is used.}
  \item Did you discuss whether the data you are using/curating contains personally identifiable information or offensive content?
    \answerNA{No such data is used.}
\end{enumerate}

\item If you used crowdsourcing or conducted research with human subjects...
\begin{enumerate}
  \item Did you include the full text of instructions given to participants and screenshots, if applicable?
    \answerNA{Neither crowdsourcing nor research on human subjects was conducted.}
  \item Did you describe any potential participant risks, with links to Institutional Review Board (IRB) approvals, if applicable?
    \answerNA{Neither crowdsourcing nor research on human subjects was conducted.}
  \item Did you include the estimated hourly wage paid to participants and the total amount spent on participant compensation?
    \answerNA{Neither crowdsourcing nor research on human subjects was conducted.}
\end{enumerate}

\end{enumerate}
\newpage
\appendix
\section{Appendix}
\subsection{Proof of Equivariance} \label{app:proof-of-equivariance}
\paragraph{SPHC Initialization}
Here we give proof that certain quantities from the main text are invariant or equivariant,  respectively. Let us start with the spherical harmonic coordinates (SPHC) which are initialized as 
\begin{align}
 \sphc_i^{(l)}(R) =  \frac{1}{\mathcal{C}_i} \sum_{j \in \neigh{i}} \frcut(r_{ij}) \cdot \sphvec{l}(\nbr_{ij})\,, \label{app:eq:sphc-initialization}
\end{align}
where $\sphvec{l}: S^2 \mapsto \RN{2l+1}$. In contrast to the main text, we make the dependence of the right hand side of \eq\eqref{app:eq:sphc-initialization} on the atomic positions $\mathcal{R} = \{\br_1, \dots, \br_n\}$ explicit. Each degree-wise entry in the initialized SPHCs (\eq\eqref{app:eq:sphc-initialization}) transforms under rotation as
\begin{align}
\begin{split}
    \sphc_i^{(l)}(\mrot\,\mathcal{R}) &\sim  \sum_{j \in \neigh{i}} \frcut(r_{ij}) \cdot \sphvec{l}(\mrot\nbr_{ij})\\
    &=  \sum_{j \in \neigh{i}} \frcut(r_{ij}) \cdot \bm{D}^l(\mrot)\, \sphvec{l}(\nbr_{ij})\\
    &= \bm{D}^l(\mrot)\,\sum_{j \in \neigh{i}} \frcut(r_{ij}) \cdot \sphvec{l}(\nbr_{ij}) \\
    & = \bm{D}^l(\mrot) \, \sphc_i^{(l)}(\mathcal{R})\,,
\end{split}
\end{align}
where $\bm{D}(\mrot)^l \in \RN{(2l+1) \times (2l+1)}$ is the Wigner-D matrix for degree $l$ given rotation $\mrot$ and $\mrot\mathcal{R} = \{\mrot\br_1, \dots, \mrot\br_n\}$. Since the cutoff function $\frcut$ takes the inter-atomic distance $r_{ij} = \lVert\br_j - \br_i\rVert_2$ as its input argument it is always invariant under rotation. Thus, each degree-wise entry $\sphc^{(l)}$ is equivariant after the initialization. 
\paragraph{SPHC Message Passing Update}
After initialization, each per-degree entry in $\sphc$ is updated as
\begin{align}
    \sphc_i^{\prime\,(l)}(\mathcal{R}) = \sphc_i^{(l)}(\mathcal{R}) + \sum_{j \in \neigh{i}} \frcut(r_{ij}) \cdot \alpha^{(l)}_{ij} \cdot  \sphvec{l}(\nbr_{ij})\,, \label{app:eq:sphc-mp}
\end{align}
where $\alpha^{(l)}_{ij}$ are rotationally invariant, per-degree self-attention coefficients. In the first layer, the first part in \eq\eqref{app:eq:sphc-mp} corresponds to the initialized SPHC entries, which have been already shown to be equivariant (see above). The second part of the equation has the same structural form as the initialization, with the additional self-attention value, which is a rotationally invariant scalar. Thus we can write
\begin{align}
\begin{split}
    \sphc_i^{\prime\,(l)}(\mrot\,\mathcal{R}) &= \bm{D}^l(\mrot)\, \sphc^{(l)}(\mathcal{R}) + \sum_{j \in \neigh{i}} \frcut(r_{ij}) \cdot \alpha^{(l)}_{ij} \cdot  \sphvec{l}(\mrot \nbr_{ij})\\
    &=\bm{D}^l(\mrot)\, \sphc^{(l)}(\mathcal{R}) + \bm{D}^l(\mrot)\, \sum_{j \in \neigh{i}} \frcut(r_{ij}) \cdot \alpha^{(l)}_{ij} \cdot  \sphvec{l}( \nbr_{ij})\\
    &=\bm{D}^l(\mrot)\, \bigg(\sphc_i^{(l)}(\mathcal{R}) + \sum_{j \in \neigh{i}} \frcut(r_{ij}) \cdot \alpha^{(l)}_{ij} \cdot  \sphvec{l}(\nbr_{ij}) \bigg)\\
    &=\bm{D}^l(\mrot)\,\sphc_i^{\prime\,(l)}(\mathcal{R})\,,
\end{split}
\end{align}
which shows that the updated $\sphc^{\prime\,(l)}$ is also equivariant. The proof of equivariance for later layers follows analogously.
\paragraph{Transformation of the SPHCs}
After having shown that each per-degree entry of $\sphc$ transforms under rotation according to the corresponding Wigner-D matrix $\mathbf{D}^l(\mrot) \in \RN{(2l+1) \times (2l+1)}$, one can write the direct sum of all Wigner-D matrices as concatenation of matrices along the diagonal
\begin{align}
\bm{D}(\mrot) = \bigoplus_{l\in\mathcal{L}} \bm{D}^l(\mrot)\,,
\end{align}
such that $\bm{D}(\mrot) \in\RN{(\lmax - \lmin +1) \times (\lmax - \lmin +1)}$ has a block diagonal structure. The, full SPHC vectors transform under rotation as
\begin{align}
\bm{D}(\mrot) \, \sphc(\mathcal{R}) = [\bm{D}^{\lmin}(\mrot) \sphc^{(\lmin)}(\mathcal{R}), \dots, \bm{D}^{\lmax}(\mrot) \sphc^{(\lmax)}(\mathcal{R})]\,.
\end{align}
As each of the blocks along the diagonal of $\bm{D}(\mrot)$ is an orthogonal Wigner-D matrix, $\bm{D}(\mrot)$ itself is also orthogonal.
\subsubsection{Invariance of the (per-degree) Norm} \label{app:equivariance-norm}
The per degree norm is used as input to the spherical filter function $\phi_s$. As shown above, each of the per-degree entries in $\sphc$ transforms under rotation as 
\begin{align}
\sphc^{(l)}(\mrot \, \mathcal{R}) = \bm{D}^l(\mrot) \sphc^{(l)}(\mathcal{R})\,.
\end{align}
The squared norm can be expressed in terms of an inner product
\begin{align}
\begin{split}
\lVert\sphc^{(l)}(\mrot \mathcal{R})\rVert_2^2 &= \big(\sphc^{(l)}(\mrot\,\mathcal{R})\big)^T \, \sphc^{(l)}(\mrot\,\mathcal{R}) \\ &= (\bm{D}^l(\mrot)\,\sphc^{(l)}(\mathcal{R}))^T \, (\bm{D}^l(\mrot)\,\sphc^{(l)}(\mathcal{R}))\\ &= \big(\sphc^{(l)}(\mathcal{R})\big)^T \underbrace{\big(\bm{D}^l(\mrot)\big)^T\,\bm{D}^l(\mrot)}_{=\mathbb{I}} \,\sphc^{(l)}(\mathcal{R})\\ &= \lVert\sphc^{(l)}(\mathcal{R})\rVert_2^2,
\end{split}
\end{align}
where we use the orthogonality of Wigner-D matrices to show that the inner product is rotationally invariant. If $\lVert\cdot\rVert_2^2$ is invariant, so is $\lVert\cdot\rVert_2$, which completes the proof of equivariance for the degree-wise norm.

The squared norm of the full SPHC vector $\sphc$ transforms under rotation as
\begin{align}
\begin{split}
    \lVert\sphc(\mrot\,\mathcal{R})\rVert_2^2 &= \big(\sphc(\mrot \, \mathcal{R})\big)^{T} \sphc(\mrot \, \mathcal{R}) \\ 
        &= \big(\bm{D}(\mrot)\sphc(\mathcal{R})\big)^{T} \big(\bm{D}(\mrot)\sphc(\mathcal{R})\big) \\
        &= \big(\sphc(\mathcal{R})\big)^T \underbrace{\big(\bm{D}(\mrot)\big)^T\,\bm{D}(\mrot)}_{= \mathbb{I}} \,\sphc(\mathcal{R}) \\
        &= \lVert\sphc(\mathcal{R})\rVert_2^2\,,
\end{split}
\end{align}
where we used that the orthogonality of $\bm{D}^l$ results in orthogonality of $\bm{D}$ (as argued above).
\subsection{Spherical Neighborhood} \label{app:spherical-neighborhood}
Starting point for the construction of the spherical neighborhood are the SPHCs $\Sphc$ in a given layer of \textsc{So3krates}. Consequently, the distance matrix in SPHC for all atomic pairs is given as an $n\times n$ matrix $\bm{X}$ with entries $\chi_{ij} = \lVert\sphc_i - \sphc_j\rVert_2$. The idea of a spherical neighborhood is to only consider atoms that lie within a certain distance \wrt each other in SPHC space, meaning only those for which $\chi_{ij} \leq \chi_{\text{cut}}$ holds. However, compared to the inter-atomic distances in Euclidean space, for which specific knowledge e.g.\ about bond lengths or interaction length scales exits, this is not the case for the rather abstract space of SPHCs. Thus, we apply a \textsc{softmax} function along each row (neighborhood) of $\bm{X}$, which gives a rescaled version of the spherical distance matrix $\tilde{\bm{X}}$ with rescaled entries $\tilde{\chi}_{ij} \in (0,1)$. Neighborhoods for each atom are then selected based on a polynomial cutoff function~\cite{klicpera2020directional}
\begin{align}
    \phi_{\chi_{\text{cut}}}(x)=1-\frac{(p+1)(p+2)}{2} x^{p}+p(p+2) x^{p+1}-\frac{p(p+1)}{2} x^{p+2},
\end{align}
where the input is given as $x = \tilde{\chi}_{ij} / \chi_{\text{cut}}$. Here we chose $\chi_{\text{cut}} = \kappa/n$, where $n$ is the number of atoms in the system and $\kappa$ allows to increase or decrease the size of the cutoff radius. By scaling with the inverse of the number of atoms in the system, we ensure that the relative number of neighboring atoms remains approximately constant, even when going to larger molecules. The parameter $\kappa$ allows to fine tune the relative number of atoms and can depended on the size and type of the molecule under investigation. For our cumulene experiments we chose $\kappa = 1$ and $p=6$, which reduces the number of considered interactions by $\sim 40\%$ compared to a global model. 
\subsection{Network Architecture} \label{app:network}
\begin{figure}
    \centering
    \includegraphics{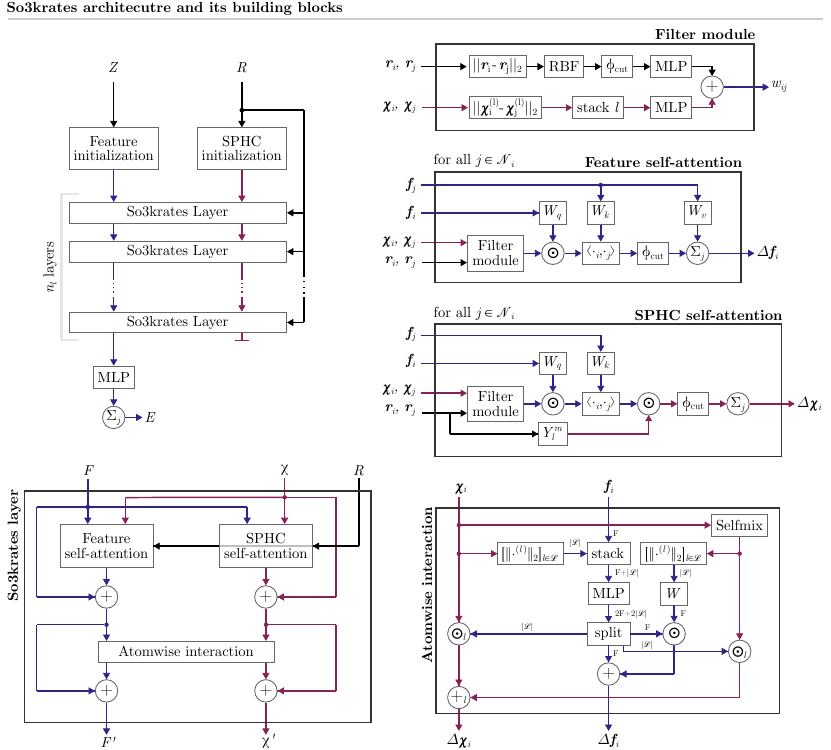}
    \caption{Figure shows the architecture of \textsc{So3krates} and its individual building blocks. More details about the different modules can be found in the corresponding sections in \ref{app:network}. Here $[\lVert\cdot_l\rVert_2]_{l \in \mathcal{L}}$ denotes per-degree norm followed by concatenation. The subscript $l$ at the pointwise product and the addition (in the atomwise interaction module) illustrates that each degree is multiplied with one entry of the $|\mathcal{L}|$-sized output vector.}
    \label{fig:architecture}
\end{figure}

The \textsc{So3krates} model is available at \url{https://github.com/thorben-frank/mlff}. It makes extensive use of the libraries \textsc{Flax}~\cite{flax2020github} for building the networks, \textsc{Optax}~\cite{optax2020github} for training and \textsc{Numpy}~\cite{harris2020array} and \textsc{Jax}~\cite{jax2018github} for additional processing steps. All code and data that is necessary to produce the results step by step presented in the main body of the paper can be downloaded from \url{https://zenodo.org/record/6584855} (newest version).
\paragraph{Detailed Network Design}
We now describe the specific implementation of the building blocks that we have formally introduced in the main text. The computational flow of the architecture is shown in \fig\ref{fig:architecture}.
\paragraph{Radial Filter}
For the radial filter function $\phi_r: \RN{} \mapsto \RN{F}$ (first summand in \eq\eqref{eq:filter}) the interatomic distances are expanded in terms of $K$ radial basis functions as proposed in \textsc{PhysNet} \cite{unke2019physnet}, which are given as
\begin{align}
    \theta^k(r_{ij}) = \frcut(r_{ij}) \,\cdot\, \exp{\big(-\gamma (\exp{(-r_{ij}}) - \mu_k)^2\big)}, \label{eq:basis-fn-expansion}
\end{align}
where $\mu_k$ is the center of the $k$-th basis function ($k = 1\dots K$). The cutoff function
\begin{align}
\frcut(r_{ij}) = \frac{1}{2}\left(\cos\left(\frac{\pi r_{ij}}{\rcut}\right) + 1\right)
\end{align}
guarantees that $\theta^k(r_{ij}) = 0$ smoothly goes to zero when interatomic distances exceed the cutoff radius $\rcut$. Here we chose $K=32$ basis functions in total. The expanded distance is passed in a two-layered MLP with \textsc{silu} non-linearity in-between and $128$ hidden and $F$ output neurons, respectively.
\paragraph{Spherical Filter}
The spherical filter function $\phi_s\left([\chi^{(l)}_{ij}]_{l\in\mathcal{L}}\right): \RN{|\mathcal{L}|} \mapsto \RN{F}$ (second summand in \eq\eqref{eq:filter}) is also modelled as a two-layered MLP with the same non-linearity. Its input is given as the stacked per-degree distances $\chi_{ij}^{(l)}$, such that the input to the MLP is of low dimension (number of spherical degrees $l$). For that reason, we only use 32 neurons in the first and again $F$ neurons in the second layer.
\paragraph{Self-Attention}
The self-attention matrix (\cf\eq\eqref{eq:self-attention}) is calculated from the filter vector $w_{ij}$ as well as from a pair of feature vectors $\bm{f}_i\in\RN{F}$ and $\bm{f}_j \in \RN{F}$. Before being passed to the inner product, each of the feature vectors is refined using a linear layer which we denote as $W_q$ and $W_k$ in the fashion of the key and query matrices usually appearing in the calculation of self-attention coefficients~\cite{vaswani2017attention}. Thus, the self-attention coefficients are calculated as 
\begin{align}
    \alpha_{ij} = \frac{1}{\sqrt{F}} \, \bm{q}_i^T (\bm{w}_{ij} \odot \bm{k}_j),
\end{align}
where $\bm{q}_i = W_q {f}_i \in \RN{F}$, $\bm{k}_j = W_k \bm{f}_j \in \RN{F}$ and $\bm{w}_{ij} \in \RN{F}$ comes from the filter module. The features $\bm{f}_j$ of the neighboring atoms are also passed through an additional linear layer $W_v$. The updated features are then given as
\begin{align}
    \bm{f}_i' = \bm{f}_i + \sum_{j \in \neigh{i}} \frcut(r_{ij})\cdot \alpha_{ij} \cdot (W_v \bm{f}_j) 
\end{align}
and the updates to the SPHCs as
\begin{align}
    \sphc_i^{\prime\,(l)} = \sphc_i^{(l)} + \sum_{j \in \neigh{i}} \frcut(r_{ij}) \cdot \alpha_{ij}^{(l)} \cdot \sphvec{l}(\br_{ij})\,.
\end{align}
Different parameters are used for the feature and SPHC updates, respectively. For the feature update, we use a predefined number of heads whereas the number of heads in the SPHC update equals the number of degrees in the SPHC vector. In order to ensure permutation invariance, all parameters of the linear layers are shared across atoms.
\paragraph{Atomwise Interaction} After the update MP step, we update the features as well as the SPHCs per atom. In this step, we do not only include cross-degree coupling in $\sphc$ but also allow for information exchange between the feature and the SPHC branch. The functions $\phi_1\left(\mathbf{f}_i, [\chi_i^{(l)}]_{l\in\mathcal{L}}, [\tilde{\chi}_i^{(l)}]_{l\in\mathcal{L}}\right)$ and $\phi_2^{(l)}\left(\mathbf{f}_i, [\chi_i^{(l)}]_{l\in\mathcal{L}}, [\tilde{\chi}_i^{(l)}]_{l\in\mathcal{L}}\right)$ are implemented by a shared MLP. The function $\phi_3^{(l)}\left([\tilde{\chi}_i^{(l)}]_{l\in\mathcal{L}}\right)$ is implemented by a single linear layer, without bias term. The full computational flow is shown in \fig\ref{fig:architecture}.
\subsection{Ablation Study Spherical Filter} \label{app:sec:ablation-study-spherical-filter}
\begin{figure}
    \centering
    \includegraphics{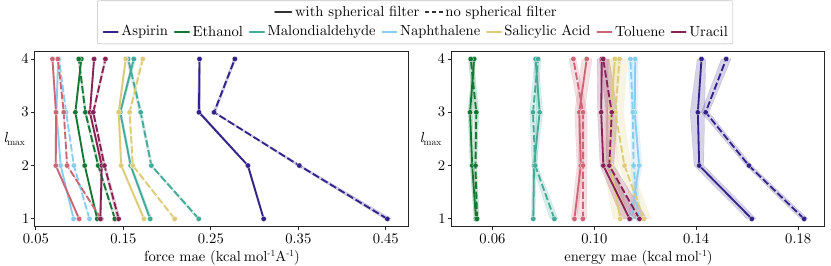}
    \caption{Ablation study on MD17 for using spherical filters. The model is evaluated on five different splits (a 1k points) of the data, where the transparent envelope is the $2\sigma$ confidence interval.}
    \label{fig:sph_filter_ablation}
\end{figure}

To illustrate the importance of the spherical filter, we examine its effect in an ablation study on the MD17 benchmark for varying maximal degree $\lmax$. As can be seen in \fig\ref{fig:sph_filter_ablation}, using spherical filters (see \eq\eqref{eq:filter}) improves performance compared to a \textsc{So3krates} model without them (solid vs. dotted lines), where the difference becomes even more pronounced for force predictions. 
Further, the effect of spherical filters becomes stronger for smaller $\lmax$. Since many equivariant MPNNs carry features including additional channels for higher order geometric information, spherical filters can be straight forwardly integrated into current architectures, offering the potential of increased accuracy.
\subsection{QM7-X Experiments} \label{app:sec:qm7x}
\begin{figure}
    \centering
    \includegraphics{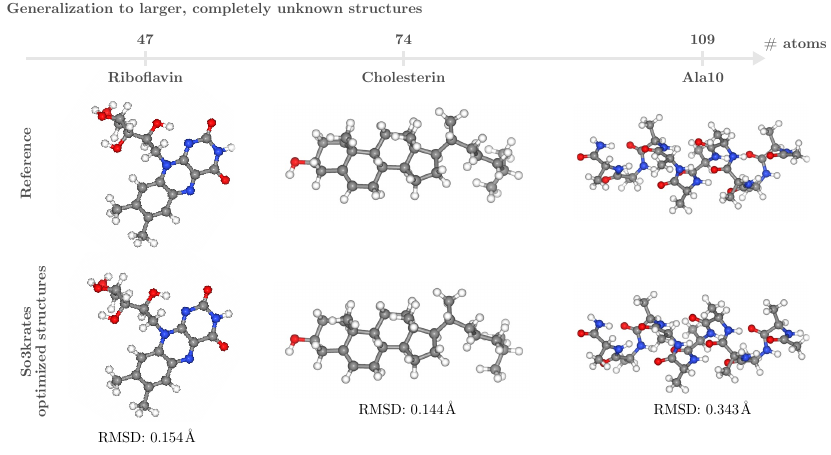}
    \caption{The upper row shows the optimized reference geometry obtained with PBE0-MBD calculation (taken from ~\cite{unke2021spookynet}). The bottom row shows geometry optimized structures obtained with a \textsc{So3krates} model \textit{with} non-local corrections from SPHC space as well as the root mean squared deviation from the reference structure. The model has been trained on the QM7-X data set for which the largest contained structure has 23 atoms. Thus, \textsc{So3krates} can generalize to larger, unknown structures without relying on the assumption of locality.}
    \label{fig:non-local-transferability}
\end{figure}

As an additional benchmark, we train \textsc{So3krates} on the full QM7-X data set and compare our results to the reported errors for known and unknown molecules in table \ref{tab:qm7x-full}. The known molecules correspond to the test samples that have not been seen during training. Following \cite{unke2021spookynet}, we use the keys (\textsc{idmol} in the QM7-X data base) \textsc{[1771, 1805, 1824, 2020, 2085, 2117, 3019, 3108, 3190, 3217, 3257, 3329, 3531, 4010, 4181, 4319, 4713, 5174, 5370, 5580, 5891, 6315, 6583, 6809, 7020]} for testing the generalization of our model to unknown structures, which have been excluded from the data set prior to generating train, test and validation splits.
\begin{table}[]
    \centering
    \resizebox{\linewidth}{!}{
    \begin{tabular}{cccccccc}
    \toprule
         & & \textsc{SchNet}~\cite{schutt2018schnet} & \textsc{PaiNN}~\cite{schutt2021equivariant} &  \textsc{SpookyNet}~\cite{unke2021spookynet} & \thead{\textsc{So3krates} \\ $F = 132$} & \thead{\textsc{So3krates} \\ $F = 264$} &  \thead{\textsc{So3krates} \\ \textit{+ non local}}
         \\\midrule
         \thead{known molecules /\\ unknown conformations} & \thead{\textit{Energy} \\ \textit{Forces}} & \thead{50.847 \\ 53.695}& \thead{15.691 \\ 20.301} & \thead{10.620 \\ 14.851} & \thead{16.815 \\ 20.422} & \thead{15.228 \\ 18.446} & \thead{16.176 \\ 19.879}
         \\\midrule
         \thead{unknown molecules /\\ unknown conformations} & \thead{\textit{Energy} \\ \textit{Forces}} & \thead{51.275 \\ 62.770} & \thead{17.594 \\ 24.161} & \thead{13.151 \\ 17.326} & \thead{21.733 \\ 25.211} & \thead{21.750 \\ 23.169} & \thead{20.071 \\ 24.236}\\
         \bottomrule
    \end{tabular}
    }
    \caption{Energy and force MAEs in meV and meV\,$\si{\angstrom}^{-1}$ for different models trained on the full QM7-X data set.}
    \label{tab:qm7x-full}
\end{table}
\paragraph{Accuracy Comparison Local vs. Non-Local Model} \label{app:sec:localvs-non-local-acc}
By comparing the results reported in table \ref{tab:qm7x-full} we find, that the non-local model shows the same accuracy as the local model for both known and unknown molecules. This underlines the applicability of the presented, non-local corrections to a large variety of molecular structures including transfer learning to unknown molecules.
\paragraph{Accuracy Improvement by Up-Scaling} \label{app:sec:upscaling}
Despite the parametric leightweight structure of \textsc{So3krates}, it is important that results can be imrproved by upscaling the model. Here we take one of the most straight forward paths and upscale the model by simply increasing the feature dimension from $F=132$ to $F = 164$. As it can be seen in table \ref{tab:qm7x-full}, this already allows to increase the accuracy on the QM7-X dataset compared to the base line model. It should be noted, however, that one could follow additional/other directions, such as increasing the maximal degree $\lmax$.
\paragraph{Generalization to Larger Molecules with a Non-Local Model} \label{sec:transfer-leraning}
As seen in table \ref{tab:qm7x-full}, a model with non-local correction is capable of generalizing to structures not seen during training. However, the re-usability of local information is one of the key assumptions for building models that can be trained on small molecules and afterwards be applied to larger, completely unknown structures~\cite{unke2021spookynet}. Thus, testing the non-local model on unknwon structures from the QM7-X data set is an insufficient test to investigate this property of transfer learning. We therefore use the non-local model for geometry optimization of molecules that range from 47 (riboflavin) up to 109 atoms (Ala10) in size (see \ref{fig:non-local-transferability}).
We find that the non-local model is robust for geometry optimization when being applied to much larger, unknown molecules. The largest molecule in the QM7-X data set has $23$ atoms.
\subsection{Time Analysis} \label{app:sec:runtime-analysis}
In order to determine the training and inference times, we follow \cite{klicpera2021gemnet} and evaluate our model on the toluene molecule from the MD17 benchmark with a batch size of 4. We trained a \textsc{NequIP} model with the same hyperparameters as reported in \cite{batzner2021se} as well as a \textsc{So3krates} model on a Tesla P100 with 12GB. The reported training time corresponds to the wall time it took each model to evaluate a single gradient update (without time for validation). We compare these runtimes to the runtimes that have been reported for \textsc{DimeNet}~\cite{klicpera2020directional} and \textsc{GemNetQ}~\cite{klicpera2021gemnet} in \cite{klicpera2021gemnet}. However, these reported times have been measured on a GeForce GTX 1080Ti (a GPU we did not have access to). In \cite{AlishaAneja} it has been found that a Tesla P100 gives a speedup factor of $\sim 1.3$, such that we downscale the reported runtimes accordingly. The resulting times are shown in table \ref{app:tab:runtime}, which are the values plotted in \ref{fig:intro-figure}d. It should be noted, that our implementation did not focus on the runtime, such that it is likely to be possible to further reduce the computational cost that is required for training and inference.
\begin{table}
    \centering
    \resizebox{\linewidth}{!}{
    \begin{tabular}{*{10}{c}}
        \toprule
        &  \multicolumn{2}{c}{\textsc{So3krates}} & \multicolumn{2}{c}{\textsc{NequIP}~\cite{batzner2021se}}
        & \multicolumn{2}{c}{\textsc{DimeNet}~\cite{klicpera2020directional}}& \multicolumn{2}{c}{\textsc{GemNetQ}~\cite{klicpera2021gemnet}}\\
        &  \textit{training} & \textit{inference} & \textit{training} & \textit{inference} & \textit{training} & \textit{inference} & \textit{training} & \textit{inference}\\\midrule
        time (ms)  & 34 & 12 & 507 & 136 & 218 & 24 & 483 & 76\\
        \bottomrule
    \end{tabular}
    }
    \caption{Training and inference times for different models for the toluene molecule from the MD17 benchmark and a batch size of 4. Hyperparemters for \textsc{So3krates} are the ones that have been used to produce the reported results on the MD17 benchmark (\cf\tab\ref{tab:MD17}). Times for \textsc{DimeNet} and \textsc{GemNetQ} have been measured on a different GPU, such that we decrease their runtimes by the factor reported here~\cite{AlishaAneja}. Inference times are for energy and force predictions.}
    \label{app:tab:runtime}
\end{table}
\subsection{Analysis of Attention Coefficients} \label{app:analysis-attention}
\begin{figure}
    \centering
    \includegraphics{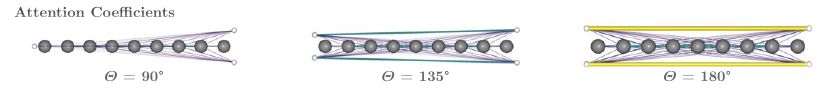}
    \caption{Visualization of the attention coefficients from the MP update of the SPHCs (\cf\eq\eqref{eq:sphc-non-local-mp}) for different dihedral angles between the rotors.}
    \label{fig:attention-analysis}
\end{figure}

In figure \ref{fig:attention-analysis} we plot the attention coefficients from the MP update of the SPHCs (\cf\eq\eqref{eq:sphc-non-local-mp}) after training for different dihedral angles between the rotors. Attention values are calculated as the average over all attention values obtained for a given atomic pair throughout all SPHC update steps. As it can be verified visually, the model picks up physically important interactions. Care should be taken, however, since it has been pointed out in \cite{ali2022xai} that looking at the bare attention values only has limited expressiveness.
\subsection{Scaling Analysis} \label{app:sec:scaling}
\begin{figure}
    \centering
    \includegraphics{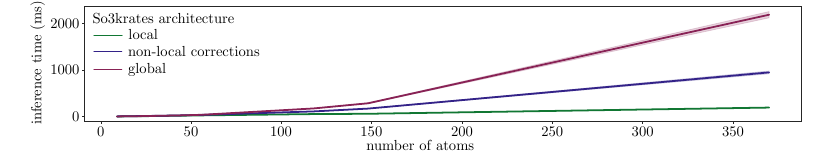}
    \caption{Scaling comparison between a local \textsc{So3krates} (as used in the first part of the experiment section \ref{sec:benchmarks-data-efficiency-generalization}.1), an architecture with non-local corrections (as used in the second part of the experiment section \ref{sec:non-local-geometric-interactions}.2) and a global \textsc{So3krates} model (only used for the sake of comparison). As expected, the local model shows linear scaling in the number of atoms and only takes $\sim 196$\,ms for a system with 370 atoms. We also find the model with non-local correction to balance between the quadratically scaling global model and the linear local model. Due to memory issues for the fully global model for larger structures, we only report inference times for energy predictions.}
    \label{fig:scaling}
\end{figure}

We further compare the scaling of three different versions of the \textsc{So3krates} model. Namely, we compare a local \textsc{So3krates} model, which only localizes in $\RN{3}$ (used in \ref{sec:benchmarks-data-efficiency-generalization}.1), a non-local \textsc{So3krates} model with non-local corrections from SPHC space (used in \ref{sec:non-local-geometric-interactions}.2) and a fully global model for which the cutoff radius is chosen such that all atoms are in each others neighborhood. The inference times for energy predictions for molecules ranging from 9 up to 370 atoms are shown in \fig\ref{fig:scaling}. As expected, we find linear scaling for the local \textsc{So3krates} version with a remarkable inference time of only $\sim 196$\,ms for 370 atoms (batch size is 25). For the fully global mode, we see quadratic scaling in the number of atoms. The model with non-local corrections can be found somewhere in-between the local and the global model. To that end, we want to stretch the fact that we did not focus on an efficient implementation for the non-local corrections. The experiments were run on a Tesla P100 with 12\,Gb and a batch size of 25.
\subsection{QM7-X250} \label{app:qm7-x250}
As starting point for the recently introduced QM7-X dataset~\cite{hoja2021qm7} serve $\sim7$k molecular graphs with a maximum of 7 non-hydrogen atoms (C, N, O, S, Cl). By sampling and optimizing structural and constitutional isomers for each graph, $\sim42$k equilibrium structures are generated. Using normal mode sampling at 1500\,K, 100 out-of-equilibrium points are generated for each structure resulting in 101 data points per structure and $\sim4.2$M geometries in total.

\begin{figure}
    \centering
    \includegraphics[width=\linewidth]{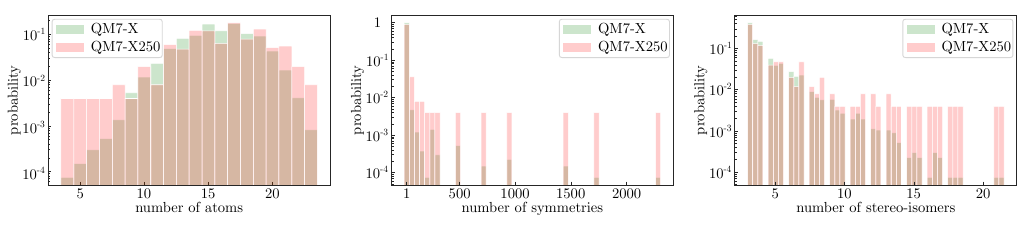}
    \caption{The figure shows the probability of occurrence for structures with a certain property value in the original QM7-X and the sub-sampled QM7-X250 dataset. As it can be seen, QM7-X follows the distribution of the original dataset. As for each property value at least once structure is present in QM7-X250 (per design), values with a low count in the original data set have higher relative importance, which can be seen e.g. for structures with only a few atoms, or with high symmetry.}
    \label{fig:qm7x250-dataset}
\end{figure}

In order to make the data set well suited for both, kernel and neural network models, we group the geometries by structural isomers which gives $\sim13$k individual data sets, each consisting of \textit{\#stereo-isomers}\,$\times101$ geometries. For each of the data sets we choose 80 points for training, 10 points for validation and the remaining points for testing after the training. Afterwards, we randomly sample 250 data sets out of the 13k data sets. The comparison of the probabilities of drawing a molecule with a given number of atoms, number of symmetries and number of stereo-isomers from the original QM7-X and the QM7-X250 data sets are shown in \fig\ref{fig:qm7x250-dataset}. Since we ensure that each of the structural subsets present in the original data set is also present in the 250 drawn samples at least once, it can be seen that these structures are over represented in the QM7-X250 data set, even though they only make up a single structure. Apart from these special structures, we see that the sampled dataset correctly reproduces the distribution of the original data set.

From the 250 sampled structures, we build 250 per structure data sets with 80 training, 10 validation and 10--3748 (depending on the number of stereo-isomers) test points, which we referred to as \textit{individual} dataset in the main text. We can further build a \textit{joint} dataset, by merging all individual structures into a single data set, which gives 20\,000 ($250 \times 80$) training, 2\,500 ($250 \times 10$) validation, and 108\,800 testing points.
\subsection{QM7-X250 Experiments} \label{app:sec:qm7x-250-experiments}
In the main text we show the force MAE as a function of maximal order $\lmax$ in \fig\ref{fig:qm7x}a. In table \ref{tab:qm7x-250}, we further show the exact errors for energy and forces as a function of maximal order $\lmax$, as well as the number of parameters per  \textsc{So3krates} model. For all models with $\lmax \geq 1$, we do not include $\lmax = 0$ within the SPHCs, since the zeroth degree evaluates to a constant one and thus does not contain any additional geometric information.
\begin{table}
    \centering
    \resizebox{\linewidth}{!}{
    \begin{tabular}{*{12}{c}}
        \toprule
        & & \multicolumn{2}{c}{\textsc{sGDML}~\cite{chmiela2019sgdml}}& \multicolumn{2}{c}{\thead{\textsc{So3krates} \\ $\lmax = 0$}} & \multicolumn{2}{c}{\thead{\textsc{So3krates} \\ $\lmax = 1$}}
        & \multicolumn{2}{c}{\thead{\textsc{So3krates} \\ $\lmax = 2$}}& \multicolumn{2}{c}{\thead{\textsc{So3krates} \\ $\lmax = 3$}} \\
        & & \textit{individual} & \textit{joint} & \textit{individual} & \textit{joint} & \textit{individual} & \textit{joint} & \textit{individual} & \textit{joint} & \textit{individual} & \textit{joint}\\\midrule
        QM7-X250 & \thead{\textit{energy}\\\textit{forces}} &\thead{67.78\\107.66} & \thead{--\\--} &  \thead{78.40\\105.80} & \thead{46.87\\57.57} & \thead{66.43\\84.54} & \thead{44.64\\52.58} & \thead{47.02\\59.33} & \thead{19.10\\27.77} & \thead{\textbf{38.40}\\\textbf{48.46}} & \thead{\textbf{17.09}\\\textbf{25.37}} \\
        \midrule
        \# parameters & & \multicolumn{2}{c}{--} & \multicolumn{2}{c}{846k} & \multicolumn{2}{c}{846k}
        & \multicolumn{2}{c}{746k} & \multicolumn{2}{c}{716k} \\
        \bottomrule
    \end{tabular}
    }
    \caption{Comparison of the averaged per structure MAE for energy (in meV) and forces (in meV\,$\si{\angstrom}^{-1}$) for \textsc{sGDML} and \textsc{So3krates} models with varying $\lmax$. Best results are in bold. Note that for small $\lmax$ the total number of parameters decreases, since in the SPHC update step, as many heads are used as there are degrees (one degree for $\lmax = 0$ and $\lmax = 1$). As a consequence, the size of the matrices $Q$ and $K$, that are applied to the feature vectors that make up the inner product, become smaller in size. The feature dimension $F = 132$ is the same for all models. This leads to a decrease of the total number of network parameters, which is larger than the increase in network parameters, due to the additional degrees. Note that it is important to take the average over per structure MAEs, since the different structures have differently many test samples (see section \ref{app:qm7-x250}).}
    \label{tab:qm7x-250}
\end{table}
The generalization to unknown molecules is tested on the same structure keys as in the \textsc{SpookyNet} paper~\cite{unke2021spookynet} and in the QM7-X experiment from above~\ref{app:sec:qm7x}.
As we only trained on forces and generalize to completely unknown molecules, we can not fit the energy integration constant (as we assume to have no reference data for the unknown molecules). For that reason we only report force errors (see \ref{tab:qm7x-generalization}).
\begin{table}
    \centering
    \begin{tabular}{*{7}{c}}
        \toprule
        & &\textsc{sGDML}~\cite{chmiela2019sgdml} & \thead{\textsc{So3krates} \\ $\lmax = 0$} & \thead{\textsc{So3krates} \\ $\lmax = 1$} & \thead{\textsc{So3krates} \\ $\lmax = 2$} & \thead{\textsc{So3krates} \\ $\lmax = 3$} \\\midrule
        Generalization & \textit{forces} & 86.84 & 159.42 & 117.91 & 76.05 & $\bm{68.29}$ \\
        \bottomrule
    \end{tabular}
    \caption{Comparison of the averaged per structure MAE for forces (in meV\,$\si{\angstrom}^{-1}$) for \textsc{sGDML} and \textsc{So3krates} models with varying $\lmax$ when applied to completely unknown structures. Best results are in bold. Due to the way the molecular descriptor is designed, \textsc{sGDML}~\cite{chmiela2019sgdml} models can only be trained on individual structures. Thus the reported results for \textsc{sGDML} are not generalization results but rather serve as a benchmark for the generalization MAE of \textsc{So3krates}. The number of parameters are the same as reported in \tab\ref{tab:qm7x-250}. Note that it is important to take the average over per structure MAEs, since the different structures have differently many test samples (see section \ref{app:qm7-x250}).}
    \label{tab:qm7x-generalization}
\end{table}
However, one could train a \textsc{So3krates} model on both, energy and forces to obtain meaningful predictions for both. In that case, atomization energies would need to be included to obtain equal energy scales across different molecules as done for the full QM7-X data set.
\subsection{Additional Experiments for Non-Local Effects} \label{app:sec:additional-experiments-non-local-effects}
\begin{figure}
    \centering
    \includegraphics{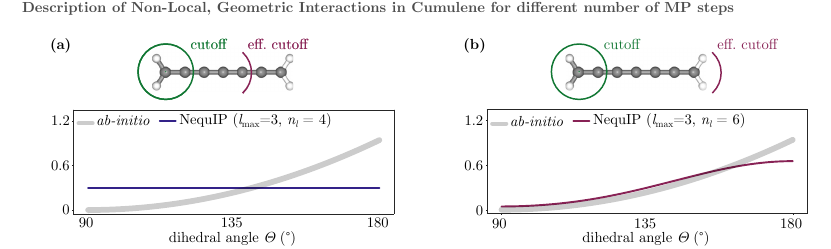}
    \caption{\textbf{(a)}~Description of the energy profile in cumulene using $\nlayer = 4$ in the \textsc{NequIP} model, such that the effective cutoff radius is smaller then the length scale on which the electronic effects take place. \textbf{(b)} By increasing the number of layers to $\nlayer = 6$, the effective cutoff becomes large enough to transfer information about the rotor orientation along the molecular graph. However, the problem is only shifted to larger length scales.}
    \label{app:fig:non-local-add}
\end{figure}

Here we demonstrate, that increasing the number of local MP steps in \textsc{NequIP} allows to model the non-local effects in cumulene, as an increase of the number of steps increases the effective cutoff of the model (see \fig\ref{app:fig:non-local-add}). However, this just shifts the problem to larger distances; increasing the length of the cumulene molecule again leads to a scenario where the local model fails. 
\subsection{Benchmarks for Non-Local Effects}
We further apply \textsc{So3krates} to the benchmark presented in \cite{ko2021fourth}, which explicitly introduces structures which exhibit non-local effects. We find, that for the carbon chain with non-local charge transfer, \textsc{So3krates} with non-local corrections improves by a factor $\sim 2$ compared to a local model. For the remaining structures no such difference can be found. Notably, the non-local effects from the presented benchmark do not result in global geometric relations but rather change the overall behavior by adding or removing certain atoms (called doping). Thus, a combination of the \textsc{So3krates} mechanism (for geometric relations) and the mechanism presented in \textsc{SpookyNet} might be a valuable direction for future work. Full results are reported in table \ref{tab:ko_at_al}. Note that \textsc{So3krates} could be easily extended to predict partial charges as well following e.g.~\cite{unke2021spookynet}.
\begin{table}[]
    \centering
    \resizebox{\linewidth}{!}{
    \begin{tabular}{cccccccc}
    \toprule
    & & \textsc{2G-BPNN}& \textsc{3G-BPNN} & \textsc{4G-BPNN} & \textsc{SpookyNet} & \textsc{So3krates} & \thead{\textsc{So3krates} \\ \textit{+ non local}}
    \\\midrule
    C$_{10}$H$_2$\,/\,C$_{10}$H$_2^+$ & \thead{\textit{Energy} \\ \textit{Forces} \\ \textit{Charges}} & \thead{1.619 \\ 129.5 \\ -- } & \thead{2.045 \\ 231.0 \\ 20.08} & \thead{1.194 \\ 78.0 \\ 6.577} & \thead{0.364 \\ 5.802 \\ 0.117} & \thead{0.113 \\ 13.198 \\ --} & \thead{0.122 \\ 7.844 \\ --}
    \\\midrule
    Na$_{8/9}$Cl$_8^+$ & \thead{\textit{Energy} \\ \textit{Forces} \\ \textit{Charges}} & \thead{1.692 \\ 57.39 \\ --} & \thead{2.042 \\76.67 \\ 20.80} & \thead{32.78 \\ 15.82 \\ 1.323} & \thead{1.052 \\ 1.052 \\ 0.111} & \thead{0.455 \\ 3.126 \\ --} & \thead{0.474 \\ 3.316 \\ --}
    \\\midrule
    Au$_2$-MgO & \thead{\textit{Energy} \\ \textit{Forces} \\ \textit{Charges}} &\thead{2.287 \\ 153.1 \\ --} & \thead{-- \\ -- \\ --} & \thead{0.219 \\ 66.0 \\ 5.698} & \thead{0.107 \\ 5.337 \\ 1.013} & \thead{0.062 \\ 8.130 \\ --} & \thead{0.064 \\ 9.472 \\ --}
    \\\bottomrule
    \end{tabular}
    }
    \caption{RSME errors for energy and forces in meV\,/\,atom and meV\,/\,$\si{\angstrom}$ for various generations of Behler-Parinello networks~\cite{ko2021fourth}, \textsc{SpookyNet}~\cite{unke2021spookynet} and a local and non-local \textsc{So3krates} model.}
    \label{tab:ko_at_al}
\end{table}
\subsection{Training Details} \label{app:sec:training-details} 
We train \textsc{So3krates} by minimizing a combined loss of energy and forces 
\begin{align}
\text{Loss} = (1 - \beta) \cdot (E - \tilde{E})^2 + \frac{\beta}{3N} \sum_{k=1}^n \sum_{i \in (x,y,z)} (F_{k}^i - \tilde{F}_k^i)^2, \label{eq:loss}
\end{align}
where $\tilde{E}$ and $\tilde{F}$ are the ground truth and $E$ and $F$ are the predictions of the model. The loss is evaluated on mini batches with a batch size given in table \tab\ref{tab:app-hyperparameters-training}. The parameter $\beta$ is used to control the trade-off between energy and forces and additionally accounts for different energy and force scales. We train our models with the \textsc{ADAM} optimizer \cite{kingma2014adam} and an initial learning rate of $\mu = 1\times10^{-3}$. We use exponential learning rate decay where the learning rate is decreased by a factor of $0.5$ every 1k epochs for the MD17 benchmark and the joint QM7-X250 dataset and every 300 epochs for individual models on the QM7-X250 dataset. For training on the full QM7-X data set we reduced the leraning rate every 250 epochs by a factor of 0.7. We further applied gradient clipping to a maximal norm of 1. 

Additional hyperparameters that have been used to produce the tables and figures in this work are given \tab\ref{tab:app-hyperparameters-training}. Whenever $\beta = 1$ is reported, no energy contribution did enter the loss function. In that case, we calculated the integration constant for energy according to section \ref{app:sec:integration-constant}.
\begin{table*}[]
    \centering
    \resizebox{\linewidth}{!}{
    \begin{tabular}{ccccccccccccc}
    \toprule
         Ref. & $F$ & $n_{\mathrm{layers}}$& $\rcut$ (\an) &  $\lmax$ & geom. corr. & spherical filter & $\beta$ & epochs & $B_s$ & $N_{\mathrm{train}}$ & $N_{\mathrm{valid}}$\\\midrule
         \fig\ref{fig:non-local-experiment} & 128 & 4 & 2.5 & 1 & True/False & True & 1 & 4k & 8 & 1k & 1k \\\midrule
         \fig\ref{fig:sphc-embedding} & 128 & 4 & 2.5 & 1 & True/False & True & 1 & 4k & 8 & 1k & 1k \\\midrule
         \fig\ref{fig:qm7x}a & 132 & 6 & 5 & [0,1,2,3] & False & True & 1 & 1.5k & 1 & 80 & 10 \\
         \midrule
         \fig\ref{fig:qm7x}b & 132 & 6 & 5 & [0,1,2,3] & False & True & 1 & 6k & 100& 20k & 2.5k \\
         \midrule
         \tab\ref{tab:MD17} & 132 & 6 & 5 & 3 & False & True & 0.99 & 4k & 8& 1k & 1k \\
         \midrule
         \fig\ref{fig:sph_filter_ablation} & [128, 128, 132, 128] & 6 & 5 & [1,2,3,4] & False & True / False & 0.99 & 4k & 8& 1k & 1k \\
         \midrule
         \fig\ref{fig:non-local-transferability} & 132 & 6 & 5 & 3 & True & True & 0.99 & 1k & 100 & 3.6M & 360k \\
         \midrule
         \tab\ref{tab:qm7x-full} & 132 & 6 & 5 & 3 & True/False & True & 0.99 & 1k & 100 & 3.6M & 360k \\
         \midrule
          \tab\ref{tab:qm7x-full} & 264 & 6 & 5 & 3 & False & True & 0.99 & 1k & 100 & 3.6M & 360k \\
           \midrule
          \tab\ref{tab:ko_at_al} & 132 & 6 & 5 & 3 & True/False & True & 0.99 & 1k & 100 & * & *\\

         \bottomrule
    \end{tabular}
    }
    \caption{\textbf{Training hyperparameters:} Table summarizes the training hyperparameters, used for the different models and experiments. Experiments are identified via their figure/table number in the main text. For the results reported in \tab\ref{tab:ko_at_al}, the same sizes of data splits as reported in \cite{ko2021fourth} have been used (indicated by an asterix).}
    \label{tab:app-hyperparameters-training}
\end{table*}
\subsection{Energy Integration Constant} \label{app:sec:integration-constant}
When only training on forces, the resulting energy predictions are likely to be shifted \wrt the correct energy values, due to vanishing constants when taking the derivative. Since force fields are conservative vector fields, one can define the following loss for the constant $c$ as
\begin{align}
\begin{split}
    \mathcal{L}(c) &= \sum_{i=1}^{M} \left( \int \mathbf{f}_{\mathrm{F}}(R_i)\,\mathrm{d}R_i - E_i \right)^2 \\
    &=\sum_{i=1}^{N_{\mathrm{data}}} \left( f_{\mathrm{E}}(R_i) + c - E_i \right)^2
\end{split}
\end{align}
where index $i$ runs over the $M$ data points, $R_i \in \RN{n \times 3}$ are the atomic coordinates and $E_i$ is the reference value of the PES. The functions $\mathbf{f}_{\mathrm{F}}: \RN{n \times 3} \mapsto \RN{n \times 3}$ and $f_{\mathrm{E}}: \RN{n \times 3} \mapsto \RN{}$ are the force and energy function, respectively. Minimization \wrt to $c$ then gives
\begin{align}
    &&\partial_c \, \mathcal{L}(c) &\overset{!}{=} 0 \\
    \iff && 2 \sum_{i=1}^{M} c - \big(E_i + f_{\mathrm{E}}(R_i)\big) &= 0 \\
    \iff && \frac{1}{M} \sum_{i=1}^M E_i + f_{\mathrm{E}}(R_i) &= c\,.
\end{align}
Thus, the shifted energy function is given as $\tilde{f}_{\mathrm{E}}(R_i) = c + f_{\mathrm{E}}(R_i)$.
\end{document}